\documentclass[10pt,twocolumn,letterpaper]{article}

\usepackage{titling}

\usepackage{cvpr}              

\usepackage{graphicx}
\usepackage{amsmath}
\usepackage{amssymb}
\usepackage{booktabs}
\usepackage{tabularx}
\usepackage[accsupp]{axessibility}

\DeclareMathOperator*{\argmax}{argmax}

\usepackage[pagebackref,breaklinks,colorlinks]{hyperref}

\usepackage[capitalize]{cleveref}
\crefname{section}{Sec.}{Secs.}
\Crefname{section}{Section}{Sections}
\Crefname{table}{Table}{Tables}
\crefname{table}{Tab.}{Tabs.}

\def\cvprPaperID{4005} 
\def\confName{CVPR}
\def\confYear{2022}

\begin{document}

\title{Ev-TTA: Test-Time Adaptation for Event-Based Object Recognition}

\author{Junho Kim$^1$, Inwoo Hwang$^1$, and Young Min Kim$^{1, 2,}$\thanks{Young Min Kim is the corresponding author.}
\\
\small
$^1$Department of Electrical and Computer Engineering, Seoul National University\\
\small
$^2$Interdisciplinary Program in Artificial Intelligence and INMC, Seoul National University\\
}
\maketitle

\begin{abstract}
We introduce Ev-TTA, a simple, effective test-time adaptation algorithm for event-based object recognition.
While event cameras are proposed to provide measurements of scenes with fast motions or drastic illumination changes, many existing event-based recognition algorithms suffer from performance deterioration under extreme conditions due to significant domain shifts.
Ev-TTA mitigates the severe domain gaps by fine-tuning the pre-trained classifiers during the test phase using loss functions inspired by the spatio-temporal characteristics of events.
Since the event data is a temporal stream of measurements, our loss function enforces similar predictions for adjacent events to quickly adapt to the changed environment online.
Also, we utilize the spatial correlations between two polarities of events to handle noise under extreme illumination, where different polarities of events exhibit distinctive noise distributions.
Ev-TTA demonstrates a large amount of performance gain on a wide range of event-based object recognition tasks without extensive additional training.
Our formulation can be successfully applied regardless of input representations and further extended into regression tasks.
We expect Ev-TTA to provide the key technique to deploy event-based vision algorithms in challenging real-world applications where significant domain shift is inevitable.

\end{abstract}

\section{Introduction}
\label{sec:intro}

\begin{figure}[t]
  \centering
  \includegraphics[width=\linewidth]{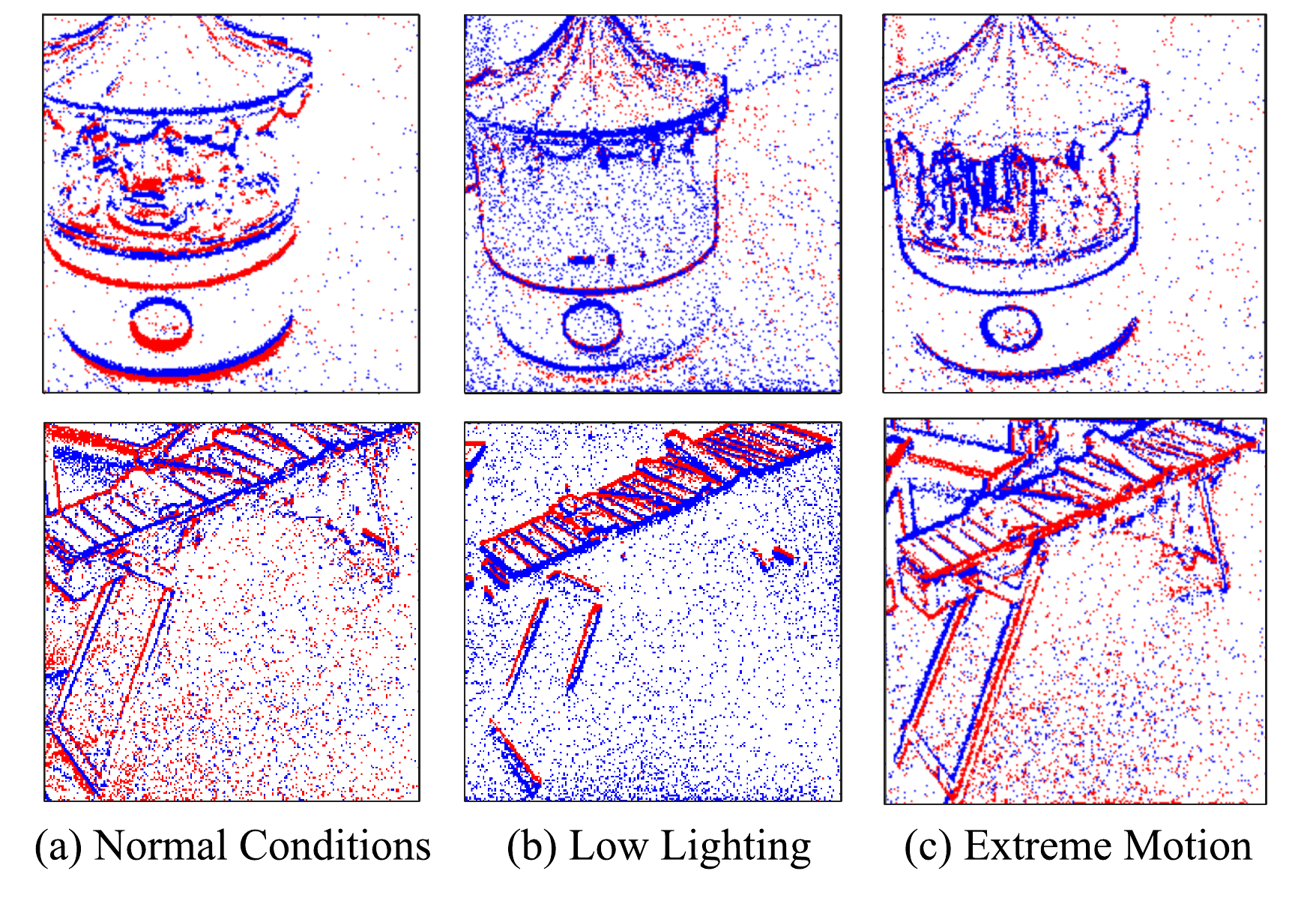}
   \caption{Visualization of events from N-ImageNet~\cite{n_imagenet} recorded in various environmental conditions. Positive, negative events are shown in blue and red, respectively. Events in low lighting (b) exhibit noise bursts, where a large number of noisy events are triggered from one polarity. Events in extreme motion (c) have denser events triggered along edges compared to normal conditions (a). Both changes lead to a significant domain gap, deteriorating the recognition performance.}
   \label{fig:change}
\vspace{-1em}
\end{figure}

Event cameras are neuromorphic sensors that produce a sequence of brightness changes with high dynamic range and microsecond-scale temporal resolution.
The sensor targets conditions where the quality of measurements degrades for standard frame-based cameras.
Conventional cameras under extreme measurement conditions produce  the prominent artifacts of motion blur or pixel saturation, and the performance deteriorates for a subsequent perceptual module.
Being able to acquire visual information in challenging environments, event cameras have the potential to overcome the limitations of frame-based cameras.
 
Despite the myriad of benefits that event cameras can offer, there is a clear gap between \emph{data acquisition} and \emph{recognition}.
While event cameras can acquire meaningful information even in challenging environments, events obtained from these conditions are typically noisy and lack visual features.
Figure~\ref{fig:change} shows that there exists a stark visual contrast between events recorded at normal lighting and regular camera motion with those from very low lighting or extreme camera motion.
Event-based object recognition algorithms are directly affected by these changes in input and the performance becomes very unstable.
Figure~\ref{fig:optim}\textcolor{red}{b} also shows the perturbation in the feature embedding space due to the domain shift.
Since it is difficult to manually collect labeled data in a wide variety of external conditions, an adaptation strategy is necessary to fully leverage the potential of event cameras.

We propose Ev-TTA, a test-time adaptation algorithm targeted for event-based object recognition.
Given a pre-trained event classifier, Ev-TTA adapts the classifier at test phase to new, unseen environments with large domain shifts.
Our method does not require labeled data from the target domain and can operate in an online manner.
Nevertheless, Ev-TTA shows a large amount of performance gain, with more than 10\% accuracy increase across all tested representations in datasets such as N-ImageNet~\cite{n_imagenet}.
While we mainly investigate domain shifts caused by external variations in camera trajectories and scene brightness, Ev-TTA is also capable of dealing with other domain shifts such as Sim2Real gap.

Ev-TTA is composed of two key components that utilize the distinctive characteristics of event data in the space-time domain.
First of all, our test-time adaptation strategy enforces the consistency of the predictions for temporally adjacent streams.
Our novel loss function jointly minimizes the discrepancy between pairs of adjacent event fragments while selectively minimizing the entropy of the predictions.
Secondly, we propose to remove events that lack spatially neighboring events in the opposite polarity.
This is based on the observation that under extreme lighting, severe noise in the event streams is exclusively generated on one polarity, as shown in Figure~\ref{fig:change}.

Since Ev-TTA only intervenes with the input event and output probability distribution, it is versatile to various event representations, datasets, or tasks.
In Section~\ref{exp:perf}, Ev-TTA shows \emph{universal} improvements across all event representations tested for a wide range of external conditions.
As there is no consensus in the optimal event representation yet, the flexibility to handle various event representations makes Ev-TTA further suitable for event data. 
Our formulation is general and is also applicable to other vision-based tasks with minor modifications.
We demonstrate that Ev-TTA could be used for tasks other than classification such as steering angle regression, suggesting the large applicability of Ev-TTA.

To summarize, our main contributions are (i) a novel test-time adaptation objective based on temporal consistency, (ii) a noise removal mechanism for low-light conditions utilizing spatial consistency, (iii) comprehensive evaluation of Ev-TTA in event-based object recognition using a wide range of event representations, and (iv) extension of Ev-TTA to event-based regression tasks.
Our experiments demonstrate that Ev-TTA can successfully adapt various event-based vision algorithms to a wide range of external conditions.

\section{Related Work}
\label{sec:related}

\paragraph{Robustness in Event-Based Object Recognition}
While event cameras can operate in harsh environments such as low-lighting and abrupt camera motion, the collected data suffer from a clear domain gap which leads to performance degradation.
Previous works have investigated the effects of motion~\cite{denoise_exp_special, hats} or night-time capture~\cite{megapixel} qualitatively or with simulated data.
Recently Deng \etal~\cite{amae} performed one of the first quantitative analyses of robustness amidst variation for a small set of motions.
Kim \etal~\cite{n_imagenet} proposed N-ImageNet along with its variants recorded under diverse camera trajectories and illumination, which enable a systematic assessment of classification robustness.
The clear performance degradation is observed for all event representations under various recording conditions.

Several event representations are hand-crafted to be robust against camera motion.
Early approaches such as event histogram~\cite{event_driving} and binary event image~\cite{binary_image_2} ignore the temporal aspects and only leverage the spatial distribution of events.
This is in contrast to other works that utilize raw timestamp  values~\cite{timestamp_image,hots,hats,evflownet}, which may be vulnerable to abrupt changes in camera speed.
To utilize the temporal information while factoring out the speed variations, several representations such as DiST~\cite{n_imagenet} and sorted time surface~\cite{ace} use relative timestamps obtained from sorting instead of absolute timestamps.

Learning-based event representations incorporate a learned module for packaging events~\cite{matrix_lstm,est}, which in theory can be trained as robust representations if provided with datasets reflecting the diverse external conditions.
However, they show competent performance only in small datasets~\cite{hats, n_caltech} and hand-crafted methods such as DiST~\cite{n_imagenet} have demonstrated performance on par with these methods in large-scale, fine-grained datasets~\cite{n_imagenet}.
This is due to the large memory requirement that inhibits large batch training, which is crucial for large-scale datasets such as N-ImageNet~\cite{n_imagenet}.

As classification algorithms based on hand-crafted representations are more often used in event-based vision~\cite{ev_gait, event_driving, evflownet, event2vid} and are sufficiently performant in large-scale datasets, we retain our focus on these class of methods.
We extensively evaluate Ev-TTA in numerous hand-crafted event representations~\cite{timestamp_image, binary_image_2, event_driving, hots, ace, n_imagenet}, and demonstrate universal performance enhancement compared to other baselines in diverse test-time conditions.

\paragraph{Test-Time Adaptation}
Unsupervised domain adaptation~\cite{sentry, uda_1, uda_2, uda_3, uda_4} aims at transferring models from a labeled source domain to an unlabeled target domain.
The objective of test-time adaptation~\cite{ttt, tta_rl, tent, tta_sr, tta_synth, tent_mum} is similar to unsupervised domain adaptation, while the difference lies in where adaptation takes place:
unsupervised domain adaptation usually undergoes an additional training phase with data from the target domain, whereas test-time adaptation mainly intervenes with the test phase.
Given the diverse changes in the input event distribution, we propose a test-time adaptation strategy reflecting the current measurement condition more adequately for practical deployment of event-based vision algorithms than collecting training datasets to capture the entire space of possible variations.

Ev-TTA takes inspiration from both unsupervised domain adaptation and test-time adaptation.
SENTRY~\cite{sentry} is one of the state-of-the-art algorithms for unsupervised domain adaptation that conditionally optimizes entropy by observing the consistency between augmented input samples.
While the training objective is effective for adaptation, SENTRY requires altering the training process and network architecture to properly function.
Tent~\cite{tent} is a lightweight approach for test-time adaptation in visual recognition, achieving large performance gain without changing the training nor network architecture. 
Tent minimizes prediction entropy during the test phase and restrains optimization to only the batch normalization layers for efficient training.
Ev-TTA leverages the strengths from both SENTRY~\cite{sentry} and Tent~\cite{tent}, while further incorporating spatio-temporal characteristics of event data for optimal performance gain.

\section{Method}
\label{sec:method}

Ev-TTA adapts a pre-trained event classifier trained on the source domain to a target domain with a significant shift in the measurement setting.
The source domain is defined as the original external condition used for training and the target domain is the new condition for testing.
For example, the classifiers could be trained with data captured in normal lighting and then tested on data under low lighting.

The raw event camera output is composed of a sequence of events, $\mathcal{E}=\{e_i=(x_i, y_i, t_i, p_i)\}$, where $e_i$ indicates brightness change with polarity $p_i \in \{-1, 1\}$ at pixel location $(x_i, y_i)$ at time $t_i$.
While there are several approaches that asynchronously process events~\cite{eventnet, asynet, snn}, we retain our focus on more prevalent approaches that employ image-like event representations.
The classification algorithms~\cite{ev_gait, event_driving, evflownet, event2vid} are composed of a two-step procedure, where events are first aggregated to form an image-like representation, and further processed with conventional image classifier architectures~\cite{resnet} to output class probabilities.

Once the input representation is chosen with the classifier $F_\theta(\cdot)$ pre-trained in the source domain, the network parameter $\theta$ for the target domain is optimized against the training objective that imposes temporal consistency between adjacent sequences of events.
The training objective is elaborated in Section~\ref{sec:objective}.

Ev-TTA can perform test-time adaptation either in an offline or online manner.
In the offline setup, Ev-TTA is first optimized for the entire target domain, and subsequently performs another set of inferences for evaluation using the same samples with the updated model parameters.
In the online setup, Ev-TTA is simultaneously evaluated and optimized, thus omitting the second inference phase.
Ev-TTA shows strong performance in both evaluation scenarios, where the detailed results are reported in Section~\ref{sec:exp}.
Note that no data from the source domain is used in training, which would lead to large amounts of additional computation as source domain data is typically much larger than the target domain.
Further, Ev-TTA does not modify the neural network architecture or the training process and thus can be applied in diverse practical settings.

The event sequence is also conditionally refined using the spatial consistency between different event polarities, and compiled into an image-like representation to serve as the input to the neural network.
The spatial consistency provides an important cue for denoising the data under extreme lighting conditions, which is further described in Section~\ref{sec:noise}.

\begin{figure}[t!]
  \centering  

   \includegraphics[width=\linewidth]{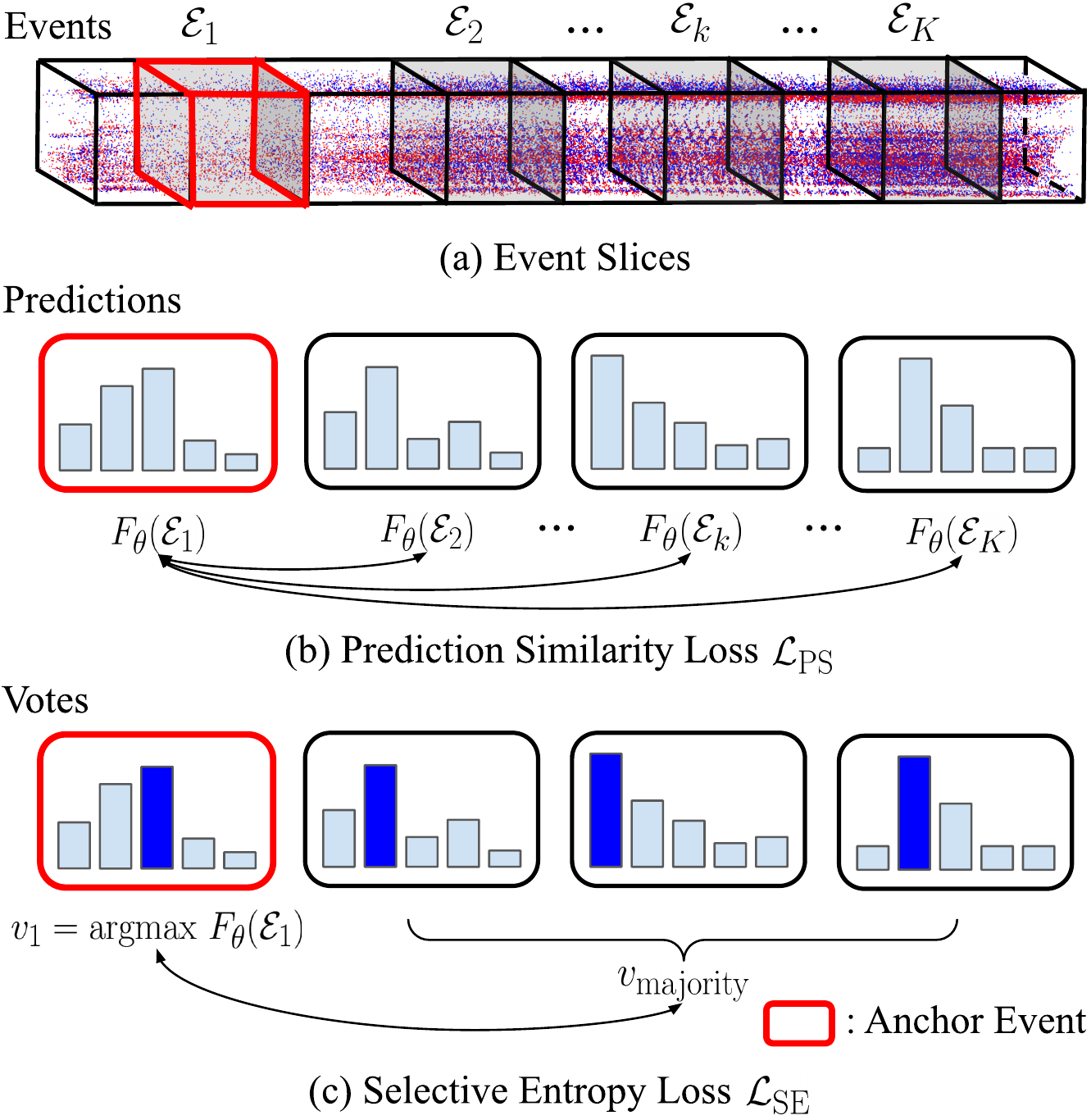}
\caption{Overview of the training objective. (a) Ev-TTA extracts $K$ random slices of equal length from the input event stream, and  fine-tunes a pre-trained classifier to enforce temporal consistency with the anchor event $\mathcal{E}_1$ and other event slices $\mathcal{E}_k$. (b) The prediction similarity loss  $\mathcal{L_\text{PS}}$ minimizes the discrepancy with respect to the anchor event (c) while the selective entropy loss $\mathcal{L_\text{SE}}$ minimizes the entropy of the anchor prediction when the votes are consistent.}
   \label{fig:overview}
\vspace{-1em}
\end{figure}


\subsection{Training Objective for Temporal Consistency}
\label{sec:objective}

Ev-TTA minimizes a loss function that imposes consistency in the time domain. 
Given an event stream $\mathcal{E}$, let $\mathcal{E}_{1}, \dots, \mathcal{E}_{K} \subset \mathcal{E}$ be the $K$ random slices of equal length obtained from $\mathcal{E}$.
Note that event-based object recognition often employs input events that span no more than 100ms~\cite{n_imagenet, n_caltech, hots}, and thus we can assume the $K$ random event slices to be temporally adjacent.
The training objective enforces the consistency between the network outputs of the event slices $F_\theta(\mathcal{E}_i), i=1,\ldots,K$, as shown in Figure~\ref{fig:overview}.
The loss function is defined as $\mathcal{L}=\mathcal{L}_{\text{PS}}+\mathcal{L}_\text{SE}$, where $\mathcal{L}_{\text{PS}}$ is the prediction similarity loss and $\mathcal{L}_\text{SE}$ is the selective entropy loss.

\paragraph{Prediction Similarity Loss}

Prediction similarity loss enforces the predicted label distributions for the temporally neighboring events $\mathcal{E}_1, \dots, \mathcal{E}_K$ to be similar, which is depicted in  Figure~\ref{fig:overview}\textcolor{red}{b}.
Using the symmetric KL divergence $S_{\text{KL}}(P, Q) = D_{\text{KL}}(P \| Q) + D_{\text{KL}}(Q \| P)$, prediction similarity loss is defined as follows,
\begin{equation}
\label{eq:sim}
    \mathcal{L}_{\text{PS}}= \frac{1}{2} \sum_{k=2}^{K} S_{\text{KL}}(F_\theta(\mathcal{E}_1), F_\theta(\mathcal{E}_k)).
\end{equation}
Note that the loss minimizes the discrepancy between the prediction for the first event slice and the rest instead of incorporating all possible pairs within the $K$ event slices.
Since the extensive pair-wise comparison would lead to a quadratic increase in computation, we instead use the first event slice as an \emph{anchor} that pulls the predictions of other event slices. 
We empirically show that using only a single event slice as an anchor is sufficient for successful adaptation, especially when it is paired with the selective entropy loss $\mathcal{L}_\text{SE}$.
We also find that the choice of the anchor does not have a significant effect on performance, where in-depth analysis is deferred to the supplementary material.

\paragraph{Selective Entropy Loss}
While the prediction similarity loss provides a meaningful learning signal for test-time adaptation, the loss heavily depends on the quality of the anchor prediction.
To this end, Ev-TTA additionally imposes the selective entropy loss $\mathcal{L}_\text{SE}$.
Inspired from SENTRY~\cite{sentry}, we propose to selectively minimize the prediction entropy of the first event slice $\mathcal{E}_{1} \subset \mathcal{E}$ only if the prediction is consistent with other event slices.
The consistency is determined by examining whether the predicted class labels are in agreement with the temporally neighboring events, as described in Figure~\ref{fig:overview}\textcolor{red}{c}.
To elaborate, each event slice $\mathcal{E}_{i}$ casts a vote on the class label with the highest probability, namely $v_{i} = \argmax F_\theta(\mathcal{E}_{i})$.
An anchor is considered consistent if its label vote $v_1$ is equal to the majority vote $v_\text{majority}$ from the other event slices $\mathcal{V}_{\text{other}}=\{v_{2}, \dots, v_{K}\}$.
Using the entropy $H(p)=-\sum_{i} p_{i} \log p_{i}$ defined for a discrete probability distribution $p \in \mathbb{R}^{{C}}$ where $C$ is number of classes, selective entropy loss is defined as follows,
\begin{equation}
     \mathcal{L}_\text{SE} = 
\left\{
	\begin{array}{ll}
		H(F_\theta(\mathcal{E}_{1}))  & \mbox{if consistent} \\
		0 & \mbox{if inconsistent.}
	\end{array}
\right.
\label{eq:select}
\end{equation}

Our loss formulation differs from the selective entropy loss of SENTRY~\cite{sentry} in two aspects.
First, the criterion for consistency is determined using temporally neighboring events, unlike the image augmentations used in SENTRY.
Further, while SENTRY~\cite{sentry} proposes to maximize the predicted entropy for samples that are inconsistent, we find that simply ignoring these samples as in Equation~\ref{eq:select} is more effective for test-time adaptation in event vision.
We further validate this claim in the ablation study in Section~\ref{exp:abl}.

\begin{figure}[t]
  \centering
   \includegraphics[width=0.9\linewidth]{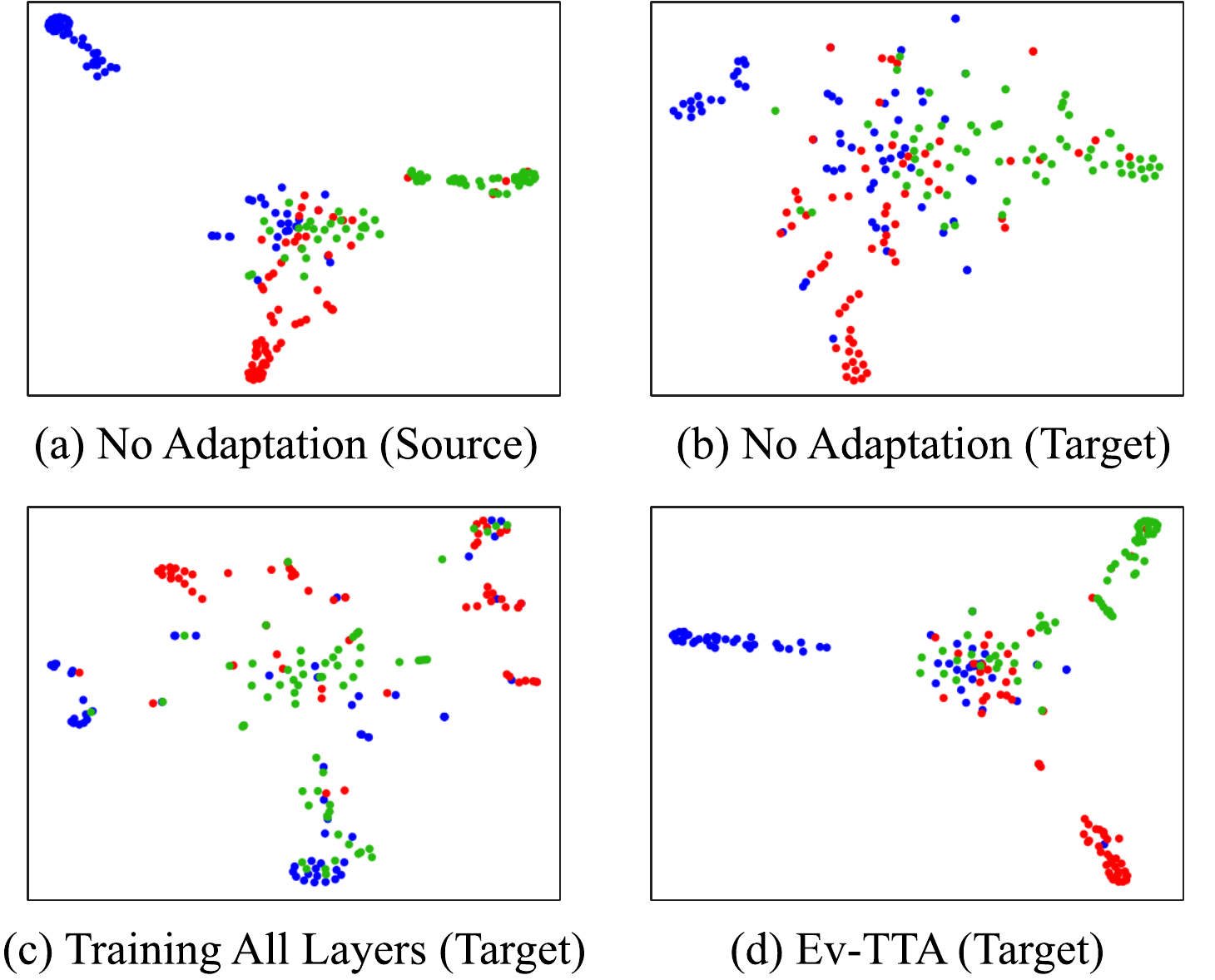}

   \caption{t-SNE~\cite{tsne} visualizations for a 3-way event classification task from N-ImageNet~\cite{n_imagenet}, trained with data captured in a normal condition and adapted to a  variant recorded under extreme camera motion. We delineate the predictions made with each adaptation method in colored circles, where each color corresponds to a label. 
   Even if the classifier is successful in the trained source domain (a), the performance does not transfer to the target domain without adequate adaptation (b).
   Training all layers fails to adapt in target data (c) as the crucial priors for event data is lost. 
    On the other hand, Ev-TTA (d) successfully adapts to target data and alleviates the performance degradation.
   }
   \label{fig:optim}
\vspace{-1em}
\end{figure}

\paragraph{Optimization Strategy}
Given the total training loss function $\mathcal{L}$, we constrain the optimization to only operate on the batch normalization layers of the pre-trained classifier as suggested by~\cite{tent}.
When the target domain data is scarce, altering the entire set of parameters may divert the model from essential priors obtained from the pre-training.
The argument is also supported in our experiment conducted with variants of N-ImageNet~\cite{n_imagenet} shown in Figure~\ref{fig:optim}.
Even using the identical objective, training the entire network results in the predicted labels to collapse (Figure~\ref{fig:optim}\textcolor{red}{c}), whereas different labels are better separated when only the batch normalization layers are optimized (Figure~\ref{fig:optim}\textcolor{red}{d}).
Ev-TTA effectively leverages the loss function that reflects the distinctive characteristics of event data and performs fast and successful adaptation, which is further discussed in Section~\ref{sec:exp}.

\paragraph{Extension to Regression}
We demonstrate that Ev-TTA could be utilized for regression, which together with classification constitute a large portion of computer vision tasks.
As a typical example, we show an extension to steering angle regression for autonomous driving.
The task is to predict the steering angle $\phi$ from a stream of events $\mathcal{E}$.

Since our loss formulation is composed of KL-divergence and entropy of the predictions, it can be easily extended to other tasks that output a probability distribution.
For steering angle regression, we design the regressor to predict both the mean and variance of the steering angle, namely $F_\theta(\mathcal{E}) = (\mu, \sigma)$.
Assuming that the output variables follow a Gaussian distribution, the regressor is trained to maximize the log likelihood as in Nix \etal~\cite{gauss},
\begin{equation}
    \mathcal{L}_\text{likelihood}=-\log \sigma - \frac{(\phi_\text{gt} - \mu)^2}{2\sigma^2},
\end{equation}
where $\phi_\text{gt}$ is the ground-truth steering angle from the source domain.

Under such conditions, we make three modifications to the loss functions used in Ev-TTA for classification.
We first replace the symmetric KL divergence from Equation~\ref{eq:sim} with the KL divergence of Gaussian distributions, namely
\begin{equation}
    S_\text{KL}(F_\theta(\mathcal{E}_1), F_\theta(\mathcal{E}_{k})) = 
    \frac{\sigma_1^4 + \sigma_{k}^4 + (\sigma_1^2+\sigma_{k}^2)(\mu_1 - \mu_{k})^2}{2\sigma_1^2\sigma_{k}^2}.
\end{equation}
We also modify the entropy from Equation~\ref{eq:select} with the entropy of Gaussian distributions, namely
\begin{equation}
    H(F_\theta(\mathcal{E}_{1})) = \log \sigma_1 \sqrt{2\pi{e}}.
\end{equation}
Finally, the consistency criterion is adapted for continuous network outputs.
An anchor event is considered consistent if its predicted variance is within a range of variances predicted from its neighbors.
To elaborate, we verify if the ratio of variances $\sigma_1^2 / \sigma_{k}^2$ for $k={2},\dots, {K}$ is bounded within $10^{-1}$ and $10$.
We impose constraints using the variance since the predicted mean may deviate largely depending on the driving scenario, whereas the predicted variance should be consistent over a longer time horizon.

With the aforementioned modifications, Ev-TTA can lead to performance enhancements in steering angle prediction, which is further discussed in Section~\ref{exp:perf}.
The result demonstrates that we can impose our adaptation strategy to other vision tasks by examining the entropy and divergence of the output distributions.

\begin{figure}[t!]
  \centering
   \includegraphics[width=\linewidth]{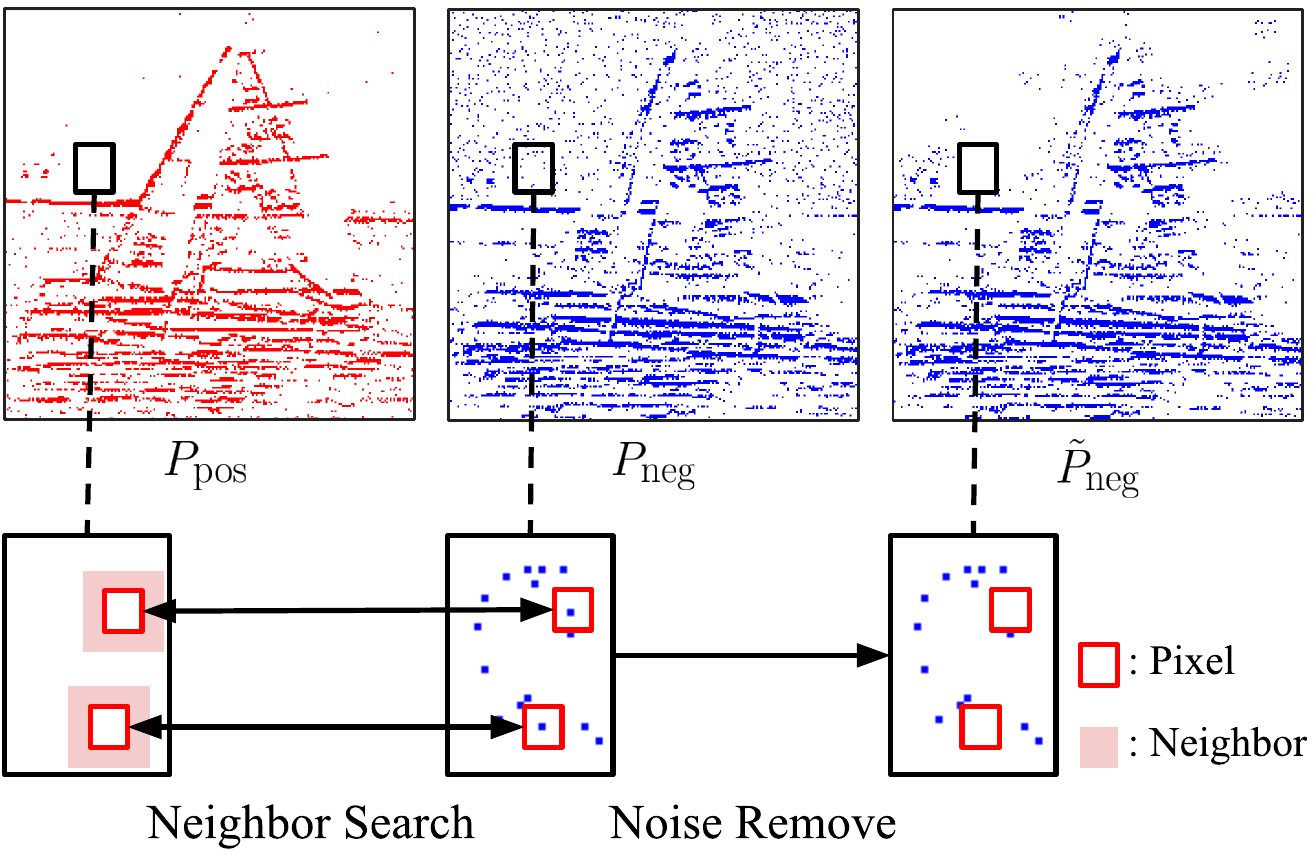}

   \caption{Illustration of conditional denoising, which is applied to events with a  large imbalance in polarity. For each pixel in the channel that contains noise burst (in this case $P_\text{neg}$), Ev-TTA first searches the spatial neighborhood in the opposite polarity. If the neighborhood lacks events, the noise is removed, and the noisy channel $P_\text{neg}$ is replaced with the denoised channel $\tilde{P}_\text{neg}$.}
   \label{fig:noise}
\vspace{-1em}
\end{figure}


\subsection{Conditional Denoising with Spatial Consistency}
\label{sec:noise}
The low light condition significantly deteriorates event-based vision algorithms, as noted by Kim \etal~\cite{n_imagenet}, and to the best of our knowledge, it has not been properly handled in previous approaches.
The main cause is the ``dark currents"~\cite{v2e}, which constantly flow through the photo-transistors.
Under low light, the currents for valid event signals become smaller, and the dark currents trigger large amounts of noise.
The severe noise in the extreme lighting condition is beyond the range of adversaries that previous approaches can handle, which are designed for small motion variation or  lighting changes~\cite{n_imagenet, ace, hats}.

We propose to conditionally remove noise in low-light conditions using a criterion derived from the spatial consistency of events.
Interestingly, we observed that the burst of noise is dominant in a single polarity, as shown in Figure~\ref{fig:change}.
We illustrate the noise removal operation using a two-channel event representation $P=\{P_\text{pos}, P_\text{neg}\} \in \mathbb{R}^{H \times W \times 2}$, where $P_\text{pos}, P_\text{neg}$ are the positive and negative channels respectively.
As shown in Figure~\ref{fig:noise}, we denoise the channel with noise burst (in this case $P_\text{neg}$) if a pixel containing events lack spatial neighbors in the opposite polarity.
The noise removal operation only takes place if there is a large imbalance in the ratio of positive and negative events.

The imbalance is formally determined with the statistical discrepancy between the positive and negative events.
Let $N_\text{pos}, N_\text{neg}$ denote the number of pixels containing positive and negative events, respectively.
Assuming $N_\text{pos}, N_\text{neg}$ follow a Gaussian distribution, the following transformation to the ratio $R=N_\text{pos} / N_\text{neg}$ follows a standard Gaussian distribution~\cite{cond_rat},
\begin{equation}
    T(R) = \frac{\mu_\text{neg}R - \mu_\text{pos}}{\sqrt{\sigma_\text{pos}^2R^2-2\rho\sigma_\text{pos}\sigma_\text{neg}R+\sigma_\text{neg}^2R^2}},
    \label{eq:transform}
\end{equation}
where $\mu_\text{pos}, \mu_\text{neg}$ are the mean, $\sigma_\text{pos}, \sigma_\text{neg}$ are the standard deviation, and $\rho$ is the cross-correlation of $N_\text{pos}, N_\text{neg}$.
To test whether the data suffers from noise burst, we transform the event ratio of the target domain using the statistics of the source domain $\{\mu_\text{pos}, \mu_\text{neg}, \sigma_\text{pos}, \sigma_\text{neg}, \rho\}$ that does not suffer from low-light conditions.
If the ratio transformed with Equation~\ref{eq:transform} follows a standard Gaussian distribution, we can assume that the target domain is free from noise burst.

The conditional denoising operation enforces spatial consistency of the two polarities on the anchor event $\mathcal{E}_1$ from Section~\ref{sec:objective}.
Given a batch of anchor events from the target domain, we compute the transformed event ratios $T(R)$ and apply statistical hypothesis testing to determine if the batch is in accordance with the source domain.
If the hypothesis test reveals that the batch contains significant polarity imbalance, we remove the detected noisy pixels based on spatial consistency, as shown in Figure~\ref{fig:noise}.
The modified channel $\Tilde{P}_\text{neg}$ replaces the original channel $P_{\text{neg}}$ to form a new anchor event representation $\Tilde{P}=\{P_\text{pos}, \Tilde{P}_\text{neg}\}$, which is subsequently used to compute the losses defined in Equation~\ref{eq:sim} and~\ref{eq:select}.
Further details about the hypothesis testing procedure are deferred to the supplementary material.

Note that our noise removal method mainly targets noise burst in low light, unlike existing denoising mechanisms~\cite{denoise_exp_special, density_filter, ev_gait} which consider a much broader set of noise.
Nevertheless, our method is extremely lightweight as it could be implemented with simple masking and effectively enhances performance, which we demonstrate in Section~\ref{exp:abl}.
\section{Experiments}
\label{sec:exp}
\begin{table*}[t]
\centering
\resizebox{0.87\linewidth}{!}{
\begin{tabular}{l|c|ccccc|cccc|c}
\toprule
Change & None & \multicolumn{5}{c|}{Trajectory} & \multicolumn{4}{c|}{Brightness} & Average \\ \midrule
Validation Dataset & Orig. & 1 & 2 & 3 & 4 & 5 & 6 & 7 & 8 & 9 & All \\ \midrule
No Adaptation & 46.76 & 43.32 & 33.78 & 39.56 & 24.78 & 36.16 & 21.52 & 30.31 & 36.60 & 34.91 & 33.44 \\
Mummadi \etal~\cite{tent_mum} & - & 46.27 & 46.04 & 46.35 & 43.27 & 44.61 & 25.59 & 35.23 & 45.73 & 45.48 & 42.07 \\
URIE~\cite{urie} & - & 42.04 & 41.45 & 42.48 & 38.66 & 40.43 & 17.59 & 29.63 & 41.77 & 41.45 & 37.28 \\
SENTRY~\cite{sentry} & - & 46.63 & 46.51 & 46.45 & 42.11 & 44.44 & 21.92 & 34.78 & 45.53 & 45.13 & 41.50 \\
Tent~\cite{tent} & - & 43.86 & 44.96 & 44.82 & 41.55 & 42.81 & 26.47 & 34.87 & 44.10 & 44.00 & 40.83 \\
Ev-TTA & - & \textbf{47.99} & \textbf{47.38} & \textbf{47.47} & \textbf{44.54} & \textbf{46.28} & \textbf{29.46} & \textbf{38.44} & \textbf{47.45} & \textbf{46.90} & \textbf{43.99} \\
\midrule\midrule
No Adaptation (Max) & - & 45.17 & 36.58 & 42.28 & 26.57 & 38.70 & 24.39 & 32.76 & 38.99 & 37.37 & 35.87 \\
Ev-TTA (Min) & - & \textbf{45.50} & \textbf{46.46} & \textbf{46.58} & \textbf{43.48} & \textbf{43.87} & \textbf{27.28} & \textbf{37.06} & \textbf{46.72} & \textbf{46.12} & \textbf{42.91} \\
\midrule\midrule
Ev-TTA (Online) & - & 44.77 & 44.80 & 45.05 & 41.77 & 43.12 & 26.43 & 35.42 & 44.42 & 44.22 & 41.11 \\
\bottomrule
\end{tabular}
}
\caption{Robustness evaluation results on N-ImageNet and its variants. The results are averaged for all tested event representations.}
\label{tab:full}
\vspace{-1em}
\end{table*}

In this section, we empirically validate various aspects of Ev-TTA.
In Section~\ref{exp:perf}, we show that the proposed test-time adaptation can enhance the performance of event-based object recognition algorithms and could be extended to steering angle prediction.  
We further validate the importance of each key constituent of Ev-TTA in Section~\ref{exp:abl}.

\paragraph{Experimental Setup}
We implement Ev-TTA using PyTorch~\cite{pytorch}, and accelerate it with an RTX 2080 GPU.
All training is performed only for one epoch, and the evaluation results are made offline unless specified otherwise.
We mostly follow the hyperparameter setup from Tent~\cite{tent}, and avoid tuning Ev-TTA as it would involve optimizing results in the test set.
Details about the hyperparameters for each dataset is deferred to the supplementary material.
Six event representations are used in the experiments: binary event image~\cite{binary_image_2}, event histogram~\cite{event_driving}, timestamp image~\cite{timestamp_image}, time surface~\cite{hots}, sorted time surface~\cite{ace}, and DiST~\cite{n_imagenet}.

\paragraph{Baselines}
The results are compared against four baseline methods: Tent~\cite{tent}, SENTRY~\cite{sentry}, Mummadi \etal~\cite{tent_mum} and URIE~\cite{urie}.
Tent~\cite{tent} and SENTRY~\cite{sentry} optimize predictions by imposing entropy minimization.
Tent optimizes only the batch normalization layers to minimize the prediction entropy.
SENTRY, on the other hand, conditionally optimizes the prediction entropy by assessing consistency from data augmentation.
We adapt SENTRY~\cite{sentry} for test-time adaptation and optimize the proposed training objective only for batch normalization layers.
The remaining two baselines focus on transforming the input representation to mitigate domain shift.
Mummadi \etal~\cite{tent_mum} propose to apply a novel input transformation network that is trained at test time to attenuate noise and other artifacts from domain shift.
URIE~\cite{urie} also proposes a similar adaptation mechanism based on input transformation networks but employs a unique attention mechanism to place more weight on salient regions in the image.
For a fair comparison with Ev-TTA, all baselines are trained during the test phase.

\subsection{Performance Enhancement}
\label{exp:perf}

\subsubsection{Event-Based Object Recognition}
\paragraph{Controlled Environments}
We first evaluate Ev-TTA using N-ImageNet~\cite{n_imagenet} to systematically evaluate the robustness enhancement under a vast range of changes.
N-ImageNet is an event-based object recognition dataset that consists of the original train set and nine variants recorded under diverse camera motion and light changes.
We train classifiers with six event representations~\cite{ace, event_driving, n_imagenet, binary_image_2, timestamp_image, hots} using the original N-ImageNet dataset, and evaluate the classifiers on the N-ImageNet variants.
Table~\ref{tab:full} displays the classification accuracy averaged across the six representations.
The large domain shift induced by these changes causes a drastic performance drop without adaptation.
Ev-TTA outperforms all other baselines and successfully adapts pre-trained classifiers to new, unseen environments.
Notably, the adapted performance is on par with the validation accuracy from the original recording, except for two variants recorded under very low lighting (dataset \# 6 and 7).
Nevertheless, a large amount of performance gain exists even in these variants, indicating the efficacy of Ev-TTA.

Further, the performance enhancement is universal, with all tested event representations showing large improvement.
This is verified by comparing `No Adaptation (Max)' from Table~\ref{tab:full}, which is the highest accuracy among the event representations for each N-ImageNet variant, with `Ev-TTA (Min)', which is the lowest accuracy for each variant.
Even the best performing representation under no adaptation is inferior to the least performing representation with Ev-TTA.
As Ev-TTA only intervenes with the input representation and the output probability distribution, it is effectively applicable to a wide range of event representations.

We further report results for the online evaluation scheme, where evaluation is performed simultaneously with training.
This reflects the practical scenario where it may not be possible to access the input data twice, and the classifier should adapt to the new environments online.
The performance of `Ev-TTA (Online)' in Table~\ref{tab:full} shows that Ev-TTA can successfully perform adaptation where large performance enhancement is universal across all tested representations.
While the offline setup provides more cues for adaptation as the data could be seen more than once, the gap between the online and offline evaluation results is not as significant.
Such results indicate that Ev-TTA can adapt both offline and online, agnostic of the underlying event representation.

\begin{table}[]
\centering
\resizebox{1.0\columnwidth}{!}{%
\begin{tabularx}{1.15\columnwidth}{l|c|c|c|c|c|c}
\toprule
Dataset & Source & Day 1 & Day 2 & Day 3 & Day 4 & Day 5\\ \midrule
None & 77.30 & 70.47 & 78.53 & 74.88 & 71.36 & 83.37 \\ 
Tent~\cite{tent} & - & 73.60 & 80.81 & 75.71 & 74.74 & 87.37 \\ 
Ev-TTA & - & \textbf{74.83} & \textbf{82.77} & \textbf{77.15} & \textbf{74.76} & \textbf{88.38} \\ \bottomrule
\end{tabularx}
}
\caption{Evaluation results on Prophesee Megapixel Dataset.}
\label{tab:prophesee}
\vspace{-0.5em}
\end{table}

\paragraph{Real-World Environments}
We also verify the adaptation of Ev-TTA in real-world recordings with uncontrolled external settings.
While N-ImageNet~\cite{n_imagenet} allows for systematic evaluation across numerous environment changes, the dataset has synthetic aspects since it is recorded with monitor displayed images.
To cope with such limitations, we test Ev-TTA on the Prophesee Megapixel dataset~\cite{megapixel}, which contains object labels for real-world recordings.
The recordings are split by day and contain five object labels from which three (car, truck, bus) are selected for the experiments.
We crop the object bounding boxes for use in classification and train a classifier on a recording from a single day, and test on five recordings from other days.
Additional details about the dataset preprocessing are provided in the supplementary material.
We compare Ev-TTA with Tent using the timestamp image~\cite{timestamp_image} representation.

As shown in Table~\ref{tab:prophesee}, Ev-TTA outperforms Tent~\cite{tent} in all tested recordings.
Compared to the plain entropy minimization of Tent~\cite{tent}, Ev-TTA imposes additional loss functions using the temporal nature of events, which leads to superior performance.
The results indicate the applicability of Ev-TTA to practical real-world scenarios incorporating event cameras.

\begin{table}[]
\centering
\resizebox{1.0\columnwidth}{!}{%
\begin{tabularx}{1.15\columnwidth}{l|c|c|c|c}
\toprule
Representation & Sim & None & Tent~\cite{tent} & Ev-TTA\\ \midrule
Timestamp Image~\cite{timestamp_image} & 53.53 & 31.36 & 38.96 & \textbf{40.66}\\ 
Binary Event Image~\cite{binary_image_2} & 54.63 & 26.62 & 38.67 & \textbf{40.94}\\ 
Event Histogram~\cite{event_driving} & 44.44 & 21.97 & 30.2 & \textbf{34.87}\\ \bottomrule
\end{tabularx}
}
\caption{Evaluation results on Sim2Real gap.}
\label{tab:sim2real}
\vspace{-1em}
\end{table}
\paragraph{Simulation and Reality Gap}
While the main focus of Ev-TTA is on adaptation amidst external changes, we demonstrate that it could also perform adaptation to reduce the simulation to reality gap.
To this end, we generate a synthetic version of N-ImageNet~\cite{n_imagenet}, termed SimN-ImageNet.
SimN-ImageNet is created with the event camera simulator Vid2E~\cite{vid2e} by moving a virtual event camera around ImageNet~\cite{imagenet} images.
Additional details about SimN-ImageNet are in the supplementary material.

We evaluate Ev-TTA for Sim2Real adaptation by applying Ev-TTA to pre-trained models in SimN-ImageNet and observing the performance change in the N-ImageNet~\cite{n_imagenet} validation set.
Table~\ref{tab:sim2real} reports the results of three tested representations, namely timestamp image~\cite{timestamp_image}, binary event image~\cite{binary_image_2}, and event histogram~\cite{event_driving}.
Ev-TTA shows the highest validation accuracy in all cases, effectively reducing the performance caused by the Sim2Real gap.
Due to the easy applicability of Ev-TTA, we expect the Sim2Real gap to be further reduced by combining Ev-TTA with recent advances in event vision for Sim2Real adaptation~\cite{reduce_sim2real, v2e, sim2real_gap_uda}.

\subsubsection{Event-Based Steering Angle Prediction}
We test our adaptation strategy into a regression task of a steering angle prediction as described in Section~\ref{sec:objective}.
We use the DDD17 dataset~\cite{ddd17}, which contains approximately 12 hours of annotated driving recordings, captured in various external conditions and organized by day.
For evaluation, we train a steering angle estimator algorithm using recordings from a single day and further evaluate the estimator on four other days.
The steering angle estimator is designed as a ResNet34~\cite{resnet} backbone receiving event histograms~\cite{event_driving} as input, following Maqueda \etal~\cite{event_driving}.

\begin{table}[]
\centering
\resizebox{1.0\columnwidth}{!}{%
\begin{tabularx}{1.15\columnwidth}{l|c|c|c|c|c}
\toprule
Scene Type & City (Source) & Freeway & City & Town & City\\ \midrule
Time & Day (Source) & Evening & Night & Day & Day\\ \midrule
None & 25.48 & 6.15 & 16.09 & 32.01 & 43.02\\ 
Tent~\cite{tent} & - & 6.52 & 15.65 & 30.94 & 41.66\\ 
Ev-TTA & - & \textbf{5.84} & \textbf{15.45} & \textbf{30.65} & \textbf{41.44}\\ \bottomrule
\end{tabularx}
}
\caption{Evaluation results on steering angle prediction using the DDD17~\cite{ddd17} dataset. The RMSE($^\circ$) is reported.}
\label{tab:ddd}
\end{table}

We report the adaptation results in Table~\ref{tab:ddd}, where the RMSE($^\circ$) with the ground-truth steering angle is measured.
Ev-TTA outperforms Tent~\cite{tent} in all tested scenarios.
By employing a subtle change in formulation, Ev-TTA could be extended to regression tasks and successfully reduce the prediction error.
However, the performance improvement is not as dramatic compared to the classification tasks.
A more effective approach for test-time adaptation in regression tasks is left as future work.

\begin{table}[]
\centering
\resizebox{0.95\columnwidth}{!}{%
\begin{tabularx}{1.15\columnwidth}{l|c|c}
\toprule
Method & Validation 6 & Validation 7\\ \midrule
Tent~\cite{tent} & 21.16 & 30.02\\ 
Tent + $\mathcal{L}_\text{PS}$ & 26.51 & 35.83\\
Tent + $\mathcal{L}_\text{PS}$ + $\mathcal{L}_\text{SE}$ & 26.82 & 36.87\\
Tent + $\mathcal{L}_\text{SE}$ (SENTRY~\cite{sentry}) & 20.13 & 33.92\\
Tent + $\mathcal{L}_\text{SE}$ (Ignore Inconsistency)& 27.13 & 36.69\\ 
Tent + $\mathcal{L}_\text{PS}$ + $\mathcal{L}_\text{SE}$ + CD (Ev-TTA) & \textbf{29.20} & \textbf{38.45}\\ \bottomrule
\end{tabularx}
}
\caption{Ablation study on the key components of Ev-TTA. $\mathcal{L}_\text{PS}$, $\mathcal{L}_\text{SE}$, CD denotes prediction similarity loss, selective entropy loss, and conditional denoising, respectively.}
\label{tab:abl}
\vspace{-1em}
\end{table}

\subsection{Ablation Study}
\label{exp:abl}

In this section, we ablate various components of Ev-TTA.
Experiments are conducted in the \# 6 and 7 variants from  N-ImageNet~\cite{n_imagenet}, using the timestamp image~\cite{timestamp_image}.
These are the most challenging splits among the N-ImageNet variants as they are recorded in low light conditions and thus contain a large amount of noise as shown in Figure~\ref{fig:change}, whose performance is also presented in Table~\ref{tab:full}.

We first examine the effect of the key constituents of Ev-TTA, namely prediction similarity loss, selective entropy loss, and conditional denoising.
As shown in Table~\ref{tab:abl}, by imposing prediction similarity loss $\mathcal{L}_\text{PS}$  on Tent~\cite{tent} (second row), a large performance enhancement takes place.
Similarly, the selective entropy loss $\mathcal{L}_\text{SE}$  also plays an important role in performance gain (third row).
Compared to SENTRY~\cite{sentry}, which maximizes entropy of inconsistent samples (fourth row),  simply ignoring such samples (Tent +  $\mathcal{L}_\text{SE}$) is much more effective (fifth row).
Finally, the conditional noise removal (CD) (Section~\ref{sec:noise}) leads to significant performance enhancement on prevalent noise bursts under low-light conditions, which can be deduced by comparing the third and sixth row of Table~\ref{tab:abl}.

\begin{figure}[t!]
  \centering
   \includegraphics[width=\linewidth]{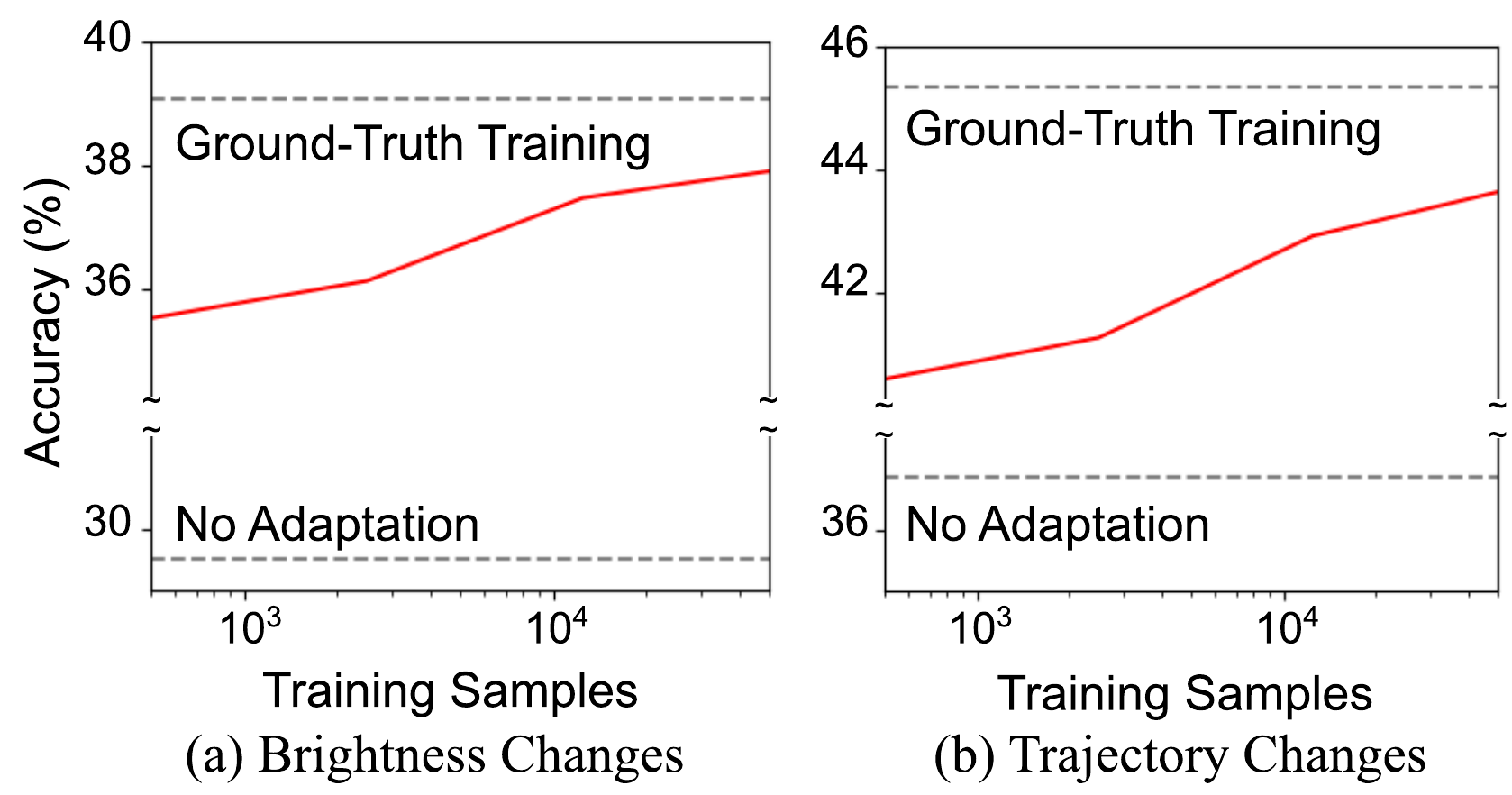}
   \caption{Effect of number of training samples on adaptation.}
   \label{fig:count}
\vspace{-1em}
\end{figure}

We further investigate the effect of the number of test-time training samples.
The six representations from Table~\ref{tab:full} are trained with varying numbers of samples and evaluated on all variants of the N-ImageNet dataset~\cite{n_imagenet}.
Figure~\ref{fig:count} shows the evaluation accuracy averaged across all representations, where the results are split by N-ImageNet variants with brightness and trajectory changes.
We additionally delineate the upper bound in performance by performing training with ground-truth labels for one epoch using the same number of training samples.
As the number of training samples increases, the average accuracy approaches the upper bound.
Furthermore, even with a very small set ($\sim500$ samples) of training data, large performance enhancement from `No Adaptation' is observable.
This demonstrates the practicality of Ev-TTA, as it can adapt in novel environments with only a small number of training data.

\section{Conclusion}
In this paper, we present Ev-TTA, a simple, effective test-time adaptation algorithm for event-based object recognition.
To alleviate the large domain shift triggered by changes in external conditions, Ev-TTA fine-tunes the pre-trained classifiers online during test phase.
The training objective is formulated by leveraging the temporal structure of events, where Ev-TTA enforces similar predictions across temporally adjacent events.
Further, to cope with noise bursts in low-light conditions, we propose a conditional denoising algorithm that employs spatial consistency.
We also extend Ev-TTA to regression tasks, by making a subtle change in the formulation.
Ev-TTA is a lightweight test-time adaptation algorithm, where universal performance enhancement occurs across various event representations in numerous tasks.
We expect Ev-TTA to facilitate the deployment of event cameras under diverse conditions and fully exploit the technical advantages of the sensor.

\paragraph{Acknowledgements}
This research was supported by the National Research Foundation of Korea (NRF) grant funded by the Korea government(MSIT) (No. 2020R1C1C1008195), Samsung Electronics Co., Ltd, Creative-Pioneering Researchers Program through Seoul National University, and Institute of Information \& communications Technology Planning \& Evaluation (IITP) grant funded by the Korea government(MSIT) [NO.2021-0-01343, Artificial Intelligence Graduate School Program (Seoul National University) and
No.2021-0-02068, Artificial Intelligence Innovation Hub].

\clearpage

\appendix
\renewcommand\thetable{\thesection.\arabic{table}}    
\setcounter{table}{0}
\renewcommand\thefigure{\thesection.\arabic{figure}}    
\setcounter{figure}{0}

\title{Supplementary Material\\Ev-TTA: Test-Time Adaptation for Event-Based Object Recognition}

\author{Junho Kim$^1$, Inwoo Hwang$^1$, and Young Min Kim$^{1, 2,}$\thanks{Young Min Kim is the corresponding author.}
\\
\small
$^1$Department of Electrical and Computer Engineering, Seoul National University\\
\small
$^2$Interdisciplinary Program in Artificial Intelligence and INMC, Seoul National University\\
}

\maketitle

\section{Hypothesis Testing}
\label{sec:hypothesis}
\setcounter{table}{0}

We handle the extreme noise in event data under low-light conditions with conditional denoising, which considers  spatial consistency, as explained in Section 3.2.
In this section, we further elaborate the hypothesis testing procedure.
Given a batch of size $N$ containing events in the target domain, we obtain the transformed event ratios $T(R_i)$ for $i=1,2,\dots,N$ using the source domain statistics, as described in Equation (6) in the main paper.
If the transformed ratios follow a standard Gaussian distribution, we can assume that the event measurement is free from noise burst.

To this end, we first calculate the batch-wise mean $\hat{\mu}$ and standard deviation $\hat{\sigma}$ of the transformed ratios $T(R_i)$.
We then determine if the event ratio $R=N_\text{pos}/N_\text{neg}$ is either too large (noise burst in positive channel) or too small (noise burst in negative channel).
Specifically, we apply the standard one-tailed z-test procedure~\cite{mathstat}, and label the batch as containing noise burst in the positive channel if the following inequality holds,
\begin{equation}
    \Phi(\frac{\sqrt{N}|\hat{\mu}-\mu_\text{thres}|}{\hat{\sigma}}) > 0.9,
\end{equation}
where $\Phi(\cdot)$ is the cumulative distribution function (CDF) of the standard Gaussian.
Here, $\mu_\text{thres}$ is the threshold value for separating noise bursts, which we set to 0.25 in all our experiments.
However, the choice of $\mu_\text{thres}$ does not have a significant impact in performance.
Table~\ref{tab:supp_mu} verifies that the accuracy of the timestamp image~\cite{timestamp_image} in validation \#6, 7 from N-ImageNet is stable for various values of $\mu_\text{thres}$.
The criterion for determining noise burst in the negative channel is similarly defined as follows,
\begin{equation}
    \Phi(\frac{\sqrt{N}|\hat{\mu}+\mu_\text{thres}|}{\hat{\sigma}}) < 0.1,
\end{equation}
where the signs of variables in the inequality are reversed.

\begin{table}[]
\centering
\resizebox{0.7\columnwidth}{!}{
\begin{tabularx}{.85\columnwidth}{l|cccc}
\toprule
$\mu_\text{thres}$ & 0.25 & 0.5 & 0.75 & 1.00\\ \midrule
Validation 6 & 29.20 & 29.39 & 29.13 & 27.57\\
Validation 7 & 38.46 & 37.36 & 37.19 & 37.21\\\bottomrule
\end{tabularx}
}
\caption{Effect of the threshold value $\mu_\text{thres}$ used in hypothesis testing on the classifier performance in N-ImageNet~\cite{n_imagenet}.}
\label{tab:supp_mu}
\end{table}

\section{Hyperparameter Setup}
\label{sec:hyper}

In this section we report the hyperparameters used for Ev-TTA.
We mostly follow the hyperparameter setup of Tent~\cite{tent} and avoid tuning the algorithm on the test set.
For all experiments, we use the Adam optimizer~\cite{adam}.
In N-ImageNet experiments, we use a learning rate of $0.00025$ with a batch size of $64$, while 
for other datasets with smaller number of labels, we use a learning rate of $0.001$ with a batch size of $128$.
For steering angle prediction, we use a learning rate of $0.000025$ with a batch size of $64$, as larger learning rates failed to converge.
We employ the identical hyperparameter setup for baselines used throughout our experiments.
\section{Dataset Preparation}
\label{sec:dataset}
\setcounter{table}{0}

\begin{table}[]
\centering
\resizebox{0.4\columnwidth}{!}{
\begin{tabularx}{0.47\columnwidth}{l|c}
\toprule
Day &  Recording\\ \midrule
Source & \texttt{2019-06-19} \\ 
Day 1 & \texttt{2019-02-22} \\
Day 2 & \texttt{2019-06-11} \\
Day 3 & \texttt{2019-06-14} \\
Day 4 & \texttt{2019-02-21} \\
Day 5 & \texttt{2019-06-26} \\
\bottomrule

\end{tabularx}
}
\caption{Conversion table for the Prophesee Megapixel Dataset~\cite{megapixel} on days used in the main paper and the corresponding recordings.}
\label{tab:supp_prophesee}
\end{table}

\begin{table}[]
\centering
\resizebox{0.8\columnwidth}{!}{
\begin{tabularx}{0.97\columnwidth}{l|ccc}
\toprule
Day &  Scene Type& Time& Recording\\ \midrule
Source & City & Day & \texttt{rec1487779465} \\ 
Day 1 & Freeway & Evening & \texttt{rec1487608147} \\
Day 2 & City & Night & \texttt{rec1487355090} \\
Day 3 & Town & Day & \texttt{rec1487856408} \\
Day 4 & City & Day & \texttt{rec1487842276} \\
\bottomrule

\end{tabularx}
}
\caption{Conversion table for the DDD17 Dataset~\cite{ddd17} on days used in the main paper and the corresponding recordings.}
\label{tab:supp_ddd}
\end{table}

In this section we explain the preprocessing pipelines used in datasets for our experiments. 
\paragraph{Prophesee Megapixel Dataset}
For evaluating Ev-TTA in real-world environments, we use the Prophesee Megapixel Dataset~\cite{megapixel} in Section 4.1.
Due to the immense size of the dataset, we select six recordings for our experiments, where the exact filename of each recording is specified in Table~\ref{tab:supp_prophesee}.
We further use the Prophesee Automotive Dataset Toolbox~\cite{megapixel} to parse the bounding boxes and collect approximately 9000 bounding boxes for three classes (car, truck, bus).
We discard other four classes (twowheeler, pedestrian, traffic light, traffic sign) in the dataset because the object bounding boxes are often too small and the class labels are not as frequent.

\paragraph{SimN-ImageNet}
To evaluate Ev-TTA for reducing sim2real gap, we generate SimN-ImageNet, which is a simulated version of N-ImageNet~\cite{n_imagenet}.
We use the event camera simulator Vid2E~\cite{vid2e,esim} to generate synthetic events from a virtual event camera moving around images from ImageNet~\cite{imagenet}.
The event camera resolution was set to $480 \times 640$, to match the resolution of the Samsung DVS camera~\cite{gen3} used for creating N-ImageNet.
Due to the large size of ImageNet~\cite{imagenet}, generating SimN-ImageNet using Vid2E~\cite{vid2e, esim} takes approximately nine days on a configuration of eight 2080Ti GPUs.

\paragraph{DDD17 Dataset}
For assessing the extension of Ev-TTA to regression tasks, we use the DDD17 dataset~\cite{ddd17} which is a dataset targeted for steering angle prediction.
We select five recordings for our experiments, where the exact filenmae of each recording is specified in Table~\ref{tab:supp_ddd}.
We further use the preprocessing toolkit provided by the authors~\cite{ddd17, ddd20} to obtain event histograms~\cite{event_driving} from raw event data.

\section{Additional Ablation Study}
\label{sec:ablation}
\setcounter{table}{0}

\begin{table}[]
\centering
\resizebox{\columnwidth}{!}{
\begin{tabularx}{1.38\columnwidth}{l|cccc}
\toprule
Method & No Adaptation & Min Entropy & Majority Vote & Random (Ours)\\ \midrule
Accuracy & 33.37 & 43.77 & 43.74 & 43.47 \\\bottomrule
\end{tabularx}
}
\caption{Ablation study on anchor event selection. We report the average accuracy of the timestamp image~\cite{timestamp_image} on the N-ImageNet variants.}
\label{tab:supp_anchor}
\end{table}

\begin{table}[]
\centering
\resizebox{0.85\columnwidth}{!}{
\begin{tabularx}{\columnwidth}{l|ccc}
\toprule
Method & No Adaptation & Ev-TTA & Augmentation\\ \midrule
Accuracy & 33.37 & 43.47 & 41.09 \\\bottomrule
\end{tabularx}
}
\caption{Ablation study on using event slices. We report the average accuracy of the timestamp image~\cite{timestamp_image} on the N-ImageNet variants.}
\label{tab:supp_slice}
\vspace{-1em}
\end{table}

\paragraph{Anchor Event Selection} 
We report the impact of choosing the anchor event for optimizing the prediction similarity loss and selective entropy loss in Section 3.1.
Recall that in Section 3.1 we choose the anchor event as a random event slice.
To validate our design choice, we implement two additional variants of Ev-TTA where the anchor is chosen more deliberately.
The first variant (Min Entropy) uses the event slice with the smallest prediction entropy as the anchor.
The second variant (Majority Vote) uses the event slice whose predicted class label is equal to the majority vote of the $K$ event slices.
We report the average performance of the timestamp image~\cite{timestamp_image} on the N-ImageNet~\cite{n_imagenet} variants under the various anchor selection schemes.
As shown in Table~\ref{tab:supp_anchor}, only a small amount of performance gain exists from using deliberate anchor selection schemes.
Therefore, the random selection scheme suffices for successful adaptation.

\paragraph{Using Event Slices for Adaptation}
We validate the use of multiple event slices for the prediction similarity loss and selective entropy loss in Section 3.1.
To this end, we implement a variant of Ev-TTA that applies data augmentation to a single event slice, similar to SENTRY~\cite{sentry}.
Instead of enforcing consistency on predictions among event slices, this variant applies the same loss formulation among augmented events.
We employ three augmentation schemes: horizontal flipping, polarity flipping, and temporal flipping.
Horizontal flipping is where the input event is flipped along the spatial dimension horizontally, and polarity flipping is where the event polarities are inverted.
Temporal flipping is where the timestamps of the input event are reversed, similar to Tulyakov \etal~\cite{timelens}.
The performance comparison between Ev-TTA and the augmentation-based variant is made on N-ImageNet~\cite{n_imagenet} using the timestamp image~\cite{timestamp_image} as input.
As shown in Table~\ref{tab:supp_slice}, the average accuracy is higher for Ev-TTA that uses event slices to impose temporal consistency.
The design choice of using event slices instead of data augmentation leads to effective adaptation.

\section{Full Evaluation Results in N-ImageNet}
\label{sec:imagenet}
\setcounter{table}{0}

In this section, we report the full evaluation results of various event representations on N-ImageNet~\cite{n_imagenet}.
Ev-TTA shows large amount of performance improvement compared to the baselines~\cite{sentry, urie, tent_mum} in all tested representations both online and offline.
The results in Table E.1$\sim$12 are the  accuracy for six event representations, namely: binary event image~\cite{binary_image_2}, event histogram~\cite{event_driving}, timestamp image~\cite{timestamp_image}, time surface~\cite{hots}, sorted time surface~\cite{ace}, and DiST~\cite{n_imagenet}.
We provide the individual accuracy for each representation.

\newpage
\begin{table*}[t]
\centering
\resizebox{0.87\linewidth}{!}{
\begin{tabular}{l|c|ccccc|cccc|c}
\toprule
Change & None & \multicolumn{5}{c|}{Trajectory} & \multicolumn{4}{c|}{Brightness} & Average \\ \midrule
Validation Dataset & Orig. & 1 & 2 & 3 & 4 & 5 & 6 & 7 & 8 & 9 & All \\ \midrule
No Adaptation & 45.86 & 43.01 & 33.62 & 39.47 & 25.39 & 36.23 & 21.16 & 30.02 & 36.52 & 34.92 & 33.37 \\
Mummadi \etal~\cite{tent_mum} & - & 44.90 & 45.25 & 45.45 & 42.66 & 43.95 & 24.27 & 33.84 & 45.00 & 44.52 & 41.09 \\
URIE~\cite{urie} & - & 41.68 & 39.77 & 42.28 & 38.30 &  39.42 & 17.68 & 30.95 & 39.38 & 41.90 & 36.82 \\
SENTRY~\cite{sentry} & - & 45.90 & 45.10 & 45.72 & 41.93 & 43.96 & 20.06 & 33.94 & 44.87 & 44.44 & 40.66 \\
Tent~\cite{tent} & - & 42.36 & 43.93 & 43.94 & 41.01 & 41.73 & 25.21 & 34.62 & 43.40 & 42.97 & 39.91 \\
Ev-TTA & - & \textbf{47.15} & \textbf{46.94} & \textbf{46.58} & \textbf{44.03} & \textbf{45.66} & \textbf{29.20} & \textbf{38.45} & \textbf{47.12} & \textbf{46.12} & \textbf{43.47} \\
\bottomrule
\end{tabular}
}
\caption{Offline evaluation results of timestamp image~\cite{timestamp_image} on N-ImageNet~\cite{n_imagenet} and its variants.}
\label{tab:supp_full}
\end{table*}

\begin{table*}[t]
\centering
\resizebox{0.87\linewidth}{!}{
\begin{tabular}{l|c|ccccc|cccc|c}
\toprule
Change & None & \multicolumn{5}{c|}{Trajectory} & \multicolumn{4}{c|}{Brightness} & Average \\ \midrule
Validation Dataset & Orig. & 1 & 2 & 3 & 4 & 5 & 6 & 7 & 8 & 9 & All \\ \midrule
No Adaptation & 45.86 & 43.01 & 33.62 & 39.47 & 25.39 & 36.23 & 21.16 & 30.02 & 36.52 & 34.92 & 33.37 \\
Mummadi \etal~\cite{tent_mum} & - & 42.60 & 43.06 & 43.39 & 40.36 & 41.29 & 24.93 & 33.55 & 42.39 & 42.27 & 39.32 \\
URIE~\cite{urie} & - & 39.12 & 38.10 & 39.74 & 36.69 &  37.48 & 18.54 & 28.32 & 38.27 & 39.07 & 35.04 \\
SENTRY~\cite{sentry} & - & 42.22 & 42.45 & 43.21 & 39.44 & 40.96 & 20.48 & 31.38 & 41.42 & 41.71 & 38.14 \\
Tent~\cite{tent} & - & 41.30 & 42.41 & 42.55 & 39.50 & 40.26 & 24.07 & 33.21 & 41.65 & 41.34 & 38.48 \\
Ev-TTA & - & \textbf{43.86} & \textbf{43.91} & \textbf{44.33} & \textbf{41.16} & \textbf{42.45} & \textbf{25.86} & \textbf{34.78} & \textbf{43.84} & \textbf{43.37} & \textbf{40.40} \\
\bottomrule
\end{tabular}
}
\caption{Online evaluation results of timestamp image~\cite{timestamp_image} on N-ImageNet~\cite{n_imagenet} and its variants.}
\label{tab:supp_full}
\end{table*}

\begin{table*}[t]
\centering
\resizebox{0.87\linewidth}{!}{
\begin{tabular}{l|c|ccccc|cccc|c}
\toprule
Change & None & \multicolumn{5}{c|}{Trajectory} & \multicolumn{4}{c|}{Brightness} & Average \\ \midrule
Validation Dataset & Orig. & 1 & 2 & 3 & 4 & 5 & 6 & 7 & 8 & 9 & All \\ \midrule
No Adaptation & 47.73 & 43.73 & 33.72 & 37.69 & 24.56 & 35.24 & 20.89 & 29.68 & 36.33 & 34.56 & 32.93 \\
Mummadi \etal~\cite{tent_mum} & - & 46.99 & 46.38 & 45.71 & 42.92 & 44.79 & 28.26 & 36.54 & 45.35 & 45.12 & 42.45 \\
URIE~\cite{urie} & - & 45.08 & 44.36 & 44.18 & 40.40 & 42.48 & 23.71 & 34.48 & 43.77 & 42.99 & 40.16 \\
SENTRY~\cite{sentry} & - & 47.06 & 48.01 & 45.75 & 41.97 & 45.06 & 24.60 & 35.48 & 45.06 & 44.91 & 41.99 \\
Tent~\cite{tent} & - & 44.88 & 45.00 & 44.20 & 41.31 & 43.11 & 26.94 & 34.65 & 43.75 & 43.57 & 40.82 \\
Ev-TTA & - & \textbf{48.64} & \textbf{48.01} & \textbf{47.24} & \textbf{44.49} & \textbf{47.06} & \textbf{30.08} & \textbf{38.34} & \textbf{47.37} & \textbf{46.58} & \textbf{44.20} \\
\bottomrule
\end{tabular}
}
\caption{Offline evaluation results of event histogram~\cite{event_driving} on N-ImageNet~\cite{n_imagenet} and its variants.}
\label{tab:supp_full}
\end{table*}

\begin{table*}[t]
\centering
\resizebox{0.87\linewidth}{!}{
\begin{tabular}{l|c|ccccc|cccc|c}
\toprule
Change & None & \multicolumn{5}{c|}{Trajectory} & \multicolumn{4}{c|}{Brightness} & Average \\ \midrule
Validation Dataset & Orig. & 1 & 2 & 3 & 4 & 5 & 6 & 7 & 8 & 9 & All \\ \midrule
No Adaptation & 47.73 & 43.73 & 33.72 & 37.69 & 24.56 & 35.24 & 20.89 & 29.68 & 36.33 & 34.56 & 32.93 \\
Mummadi \etal~\cite{tent_mum} & - & 43.71 & 43.67 & 43.20 & 40.33 & 42.54 & 25.65 & 33.66 & 42.55 & 42.76 & 39.79 \\
URIE~\cite{urie} & - & 41.94 & 42.16 & 42.10 & 38.67 & 41.10 & 23.21 & 31.90 & 40.97 & 41.20 & 38.14 \\
SENTRY~\cite{sentry} & - & 43.31 & 42.77 & 42.78 & 39.33 & 41.68 & 23.20 & 32.36 & 41.91 & 41.86 & 38.80 \\
Tent~\cite{tent} & - & 42.69 & 42.93 & 42.56 & 39.61 & 41.79 & 25.07 & 32.83 & 41.68 & 41.82 & 39.00 \\
Ev-TTA & - & \textbf{44.94} & \textbf{44.63} & \textbf{43.31} & \textbf{41.48} & \textbf{43.46} & \textbf{26.89} & \textbf{34.71} & \textbf{43.86} & \textbf{43.42} & \textbf{40.86} \\
\bottomrule
\end{tabular}
}
\caption{Online evaluation results of event histogram~\cite{event_driving} on N-ImageNet~\cite{n_imagenet} and its variants.}
\label{tab:supp_full}
\end{table*}

\begin{table*}[t]
\centering
\resizebox{0.87\linewidth}{!}{
\begin{tabular}{l|c|ccccc|cccc|c}
\toprule
Change & None & \multicolumn{5}{c|}{Trajectory} & \multicolumn{4}{c|}{Brightness} & Average \\ \midrule
Validation Dataset & Orig. & 1 & 2 & 3 & 4 & 5 & 6 & 7 & 8 & 9 & All \\ \midrule
No Adaptation & 46.36 & 42.68 & 30.68 & 37.74 & 22.99 & 34.74 & 19.00 & 27.85 & 34.03 & 32.08 & 31.31 \\
Mummadi \etal~\cite{tent_mum} & - & 46.07 & 45.02 & 44.94 & 42.35 & 43.95 & 22.90 & 31.58 & 44.66 & 45.50 & 40.77 \\
URIE~\cite{urie} & - & 42.63 & 39.30 & 42.74 & 37.28 & 41.30 & 14.58 & 25.76 & 42.23 & 42.53 & 36.48 \\
SENTRY~\cite{sentry} & - & 46.43 & 44.27 & 44.39 & 40.20 & 43.56 & 18.54 & 31.94 & 43.69 & 43.52 & 39.62 \\
Tent~\cite{tent} & - & 43.16 & 43.51 & 43.11 & 40.47 & 42.21 & 25.33 & 33.28 & 42.91 & 43.90 & 39.76 \\
Ev-TTA & - & \textbf{48.51} & \textbf{46.46} & \textbf{47.01} & \textbf{43.48} & \textbf{47.10} & \textbf{29.08} & \textbf{38.39} & \textbf{46.72} & \textbf{46.76} & \textbf{43.72} \\
\bottomrule
\end{tabular}
}
\caption{Offline evaluation results of binary event image~\cite{binary_image_2} on N-ImageNet~\cite{n_imagenet} and its variants.}
\label{tab:supp_full}
\end{table*}

\begin{table*}[t]
\centering
\resizebox{0.87\linewidth}{!}{
\begin{tabular}{l|c|ccccc|cccc|c}
\toprule
Change & None & \multicolumn{5}{c|}{Trajectory} & \multicolumn{4}{c|}{Brightness} & Average \\ \midrule
Validation Dataset & Orig. & 1 & 2 & 3 & 4 & 5 & 6 & 7 & 8 & 9 & All \\ \midrule
No Adaptation & 46.36 & 42.68 & 30.68 & 37.74 & 22.99 & 34.74 & 19.00 & 27.85 & 34.03 & 32.08 & 31.31 \\
Mummadi \etal~\cite{tent_mum} & - & 43.61 & 42.63 & 42.65 & 40.14 & 41.80 & 23.63 & 32.45 & 42.27 & 42.77 & 39.11 \\
URIE~\cite{urie} & - & 41.20 & 39.49 & 41.15 & 37.01 & 40.01 & 19.89 & 28.12 & 40.89 & 40.54 & 36.48 \\
SENTRY~\cite{sentry} & - & 42.99 & 41.75 & 42.05 & 38.14 & 41.05 & 19.52 & 30.39 & 41.00 & 41.58 & 37.61 \\
Tent~\cite{tent} & - & 42.07 & 41.78 & 41.64 & 39.12 & 40.89 & 24.05 & 31.97 & 41.44 & 41.94 & 38.32 \\
Ev-TTA & - & \textbf{44.97} & \textbf{43.73} & \textbf{43.89} & \textbf{40.85} & \textbf{43.34} & \textbf{25.42} & \textbf{34.65} & \textbf{43.68} & \textbf{43.80} & \textbf{40.48} \\
\bottomrule
\end{tabular}
}
\caption{Online evaluation results of binary event image~\cite{binary_image_2} on N-ImageNet~\cite{n_imagenet} and its variants.}
\label{tab:supp_full}
\end{table*}

\begin{table*}[t]
\centering
\resizebox{0.87\linewidth}{!}{
\begin{tabular}{l|c|ccccc|cccc|c}
\toprule
Change & None & \multicolumn{5}{c|}{Trajectory} & \multicolumn{4}{c|}{Brightness} & Average \\ \midrule
Validation Dataset & Orig. & 1 & 2 & 3 & 4 & 5 & 6 & 7 & 8 & 9 & All \\ \midrule
No Adaptation & 44.32 & 41.01 & 34.63 & 40.00 & 25.48 & 34.89 & 22.12 & 31.27 & 37.12 & 35.36 & 33.54 \\
Mummadi \etal~\cite{tent_mum} & - & 44.40 & 44.85 & 46.56 & 43.05 & 42.96 & 24.05 & 34.18 & 45.56 & 44.76 & 41.15 \\
URIE~\cite{urie} & - & 36.21 & 38.20 & 36.76 & 34.42 & 37.85 & 10.74 & 24.44 & 38.37 & 38.25 & 32.80 \\
SENTRY~\cite{sentry} & - & 44.42 & 46.63 & 47.02 & 42.27 & 42.51 & 21.00 & 35.13 & 45.90 & 45.34 & 41.14 \\
Tent~\cite{tent} & - & 41.77 & 45.23 & 45.26 & 41.69 & 41.36 & 26.03 & 34.64 & 43.97 & 43.71 & 40.41 \\
Ev-TTA & - & \textbf{45.50} & \textbf{47.42} & \textbf{47.24} & \textbf{44.27} & \textbf{43.87} & \textbf{27.28} & \textbf{37.06} & \textbf{47.05} & \textbf{46.54} & \textbf{42.91} \\
\bottomrule
\end{tabular}
}
\caption{Offline evaluation results of time surface~\cite{hots} on N-ImageNet~\cite{n_imagenet} and its variants.}
\label{tab:supp_full}
\end{table*}

\begin{table*}[t]
\centering
\resizebox{0.87\linewidth}{!}{
\begin{tabular}{l|c|ccccc|cccc|c}
\toprule
Change & None & \multicolumn{5}{c|}{Trajectory} & \multicolumn{4}{c|}{Brightness} & Average \\ \midrule
Validation Dataset & Orig. & 1 & 2 & 3 & 4 & 5 & 6 & 7 & 8 & 9 & All \\ \midrule
No Adaptation & 44.32 & 41.01 & 34.63 & 40.00 & 25.48 & 34.89 & 22.12 & 31.27 & 37.12 & 35.36 & 33.54 \\
Mummadi \etal~\cite{tent_mum} & - & 41.03 & 44.17 & 45.01 & 41.01 & 40.43 & 25.07 & 33.97 & 43.33 & 43.28 & 39.70 \\
URIE~\cite{urie} & - & 34.24 & 34.71 & 35.11 & 30.76 & 33.13 & 12.50 & 21.88 & 33.83 & 32.59 & 29.86 \\
SENTRY~\cite{sentry} & - & 40.63 & 43.79 & 44.62 & 39.62 & 39.00 & 21.49 & 32.78 & 42.74 & 42.61 & 38.59 \\
Tent~\cite{tent} & - & 39.77 & 43.60 & 44.23 & 40.20 & 39.36 & 25.13 & 33.33 & 42.50 & 42.39 & 38.95 \\
Ev-TTA & - & \textbf{42.51} & \textbf{45.18} & \textbf{45.29} & \textbf{41.37} & \textbf{40.97} & \textbf{25.68} & \textbf{35.25} & \textbf{44.15} & \textbf{43.88} & \textbf{40.48} \\
\bottomrule
\end{tabular}
}
\caption{Online evaluation results of time surface~\cite{hots} on N-ImageNet~\cite{n_imagenet} and its variants.}
\label{tab:supp_full}
\end{table*}

\begin{table*}[t]
\centering
\resizebox{0.87\linewidth}{!}{
\begin{tabular}{l|c|ccccc|cccc|c}
\toprule
Change & None & \multicolumn{5}{c|}{Trajectory} & \multicolumn{4}{c|}{Brightness} & Average \\ \midrule
Validation Dataset & Orig. & 1 & 2 & 3 & 4 & 5 & 6 & 7 & 8 & 9 & All \\ \midrule
No Adaptation & 47.90 & 44.33 & 33.50 & 40.17 & 23.72 & 37.19 & 21.57 & 30.31 & 36.63 & 35.18 & 33.62 \\
Mummadi \etal~\cite{tent_mum} & - & 47.26 & 47.35 & 47.47 & 44.29 & 45.64 & 25.56 & 36.63 & 46.60 & 46.20 & 43.00 \\
URIE~\cite{urie} & - & 42.87 & 43.76 & 44.90 & 40.50 & 40.82 & 22.05 & 33.55 & 43.37 & 41.70 & 39.28 \\
SENTRY~\cite{sentry} & - & 47.66 & 47.32 & 47.45 & 42.55 & 45.25 & 22.04 & 34.66 & 46.12 & 45.84 & 42.10 \\
Tent~\cite{tent} & - & 44.65 & 45.94 & 45.78 & 42.10 & 43.91 & 27.12 & 35.11 & 44.96 & 44.55 & 41.57 \\
Ev-TTA & - & \textbf{49.58} & \textbf{47.67} & \textbf{48.36} & \textbf{45.59} & \textbf{46.72} & \textbf{30.07} & \textbf{39.30} & \textbf{48.24} & \textbf{47.76} & \textbf{44.81} \\
\bottomrule
\end{tabular}
}
\caption{Offline evaluation results of sorted time surface~\cite{ace} on N-ImageNet~\cite{n_imagenet} and its variants.}
\label{tab:supp_full}
\end{table*}

\begin{table*}[t]
\centering
\resizebox{0.87\linewidth}{!}{
\begin{tabular}{l|c|ccccc|cccc|c}
\toprule
Change & None & \multicolumn{5}{c|}{Trajectory} & \multicolumn{4}{c|}{Brightness} & Average \\ \midrule
Validation Dataset & Orig. & 1 & 2 & 3 & 4 & 5 & 6 & 7 & 8 & 9 & All \\ \midrule
No Adaptation & 47.90 & 44.33 & 33.50 & 40.17 & 23.72 & 37.19 & 21.57 & 30.31 & 36.63 & 35.18 & 33.62 \\
Mummadi \etal~\cite{tent_mum} & - & 44.49 & 45.03 & 45.15 & 41.68 & 42.84 & 26.07 & 35.10 & 44.15 & 44.08 & 40.95 \\
URIE~\cite{urie} & - & 40.13 & 40.60 & 41.29 & 37.30 & 39.24 & 20.80 & 30.45 & 39.54 & 40.41 & 36.64 \\
SENTRY~\cite{sentry} & - & 43.98 & 44.10 & 44.79 & 39.97 & 42.01 & 21.71 & 32.38 & 43.03 & 43.56 & 39.50 \\
Tent~\cite{tent} & - & 43.40 & 44.24 & 44.30 & 40.70 & 42.18 & 25.64 & 34.08 & 43.16 & 43.13 & 40.09 \\
Ev-TTA & - & \textbf{46.02} & \textbf{45.29} & \textbf{45.91} & \textbf{42.53} & \textbf{43.90} & \textbf{26.70} & \textbf{36.17} & \textbf{45.00} & \textbf{45.22} & \textbf{41.86} \\
\bottomrule
\end{tabular}
}
\caption{Online evaluation results of sorted time surface~\cite{ace} on N-ImageNet~\cite{n_imagenet} and its variants.}
\label{tab:supp_full}
\end{table*}

\begin{table*}[t]
\centering
\resizebox{0.87\linewidth}{!}{
\begin{tabular}{l|c|ccccc|cccc|c}
\toprule
Change & None & \multicolumn{5}{c|}{Trajectory} & \multicolumn{4}{c|}{Brightness} & Average \\ \midrule
Validation Dataset & Orig. & 1 & 2 & 3 & 4 & 5 & 6 & 7 & 8 & 9 & All \\ \midrule
No Adaptation & 48.43 & 45.17 & 36.58 & 42.28 & 26.57 & 38.70 & 24.39 & 32.76 & 38.99 & 37.37 & 35.89 \\
Mummadi \etal~\cite{tent_mum} & - & 48.02 & 47.41 & 47.98 & 44.37 & 46.39 & 28.52 & 38.60 & 47.21 & 46.80 & 43.92 \\
URIE~\cite{urie} & - & 43.78 & 43.31 & 44.03 & 41.04 & 40.71 & 16.80 & 28.61 & 43.52 & 41.33 & 38.13 \\
SENTRY~\cite{sentry} & - & 48.33 & 47.70 & 48.38 & 43.71 & 46.28 & 25.25 & 37.51 & 47.53 & 46.73 & 43.49 \\
Tent~\cite{tent} & - & 46.32 & 46.17 & 46.64 & 42.74 & 44.56 & 28.20 & 36.93 & 45.59 & 45.32 & 42.50 \\
Ev-TTA & - & \textbf{48.53} & \textbf{47.75} & \textbf{48.38} & \textbf{45.35} & \textbf{47.26} & \textbf{31.02} & \textbf{39.07} & \textbf{48.19} & \textbf{47.66} & \textbf{44.80} \\
\bottomrule
\end{tabular}
}
\caption{Offline evaluation results of DiST~\cite{n_imagenet} on N-ImageNet~\cite{n_imagenet} and its variants.}
\label{tab:supp_full}
\end{table*}

\begin{table*}[t]
\centering
\resizebox{0.87\linewidth}{!}{
\begin{tabular}{l|c|ccccc|cccc|c}
\toprule
Change & None & \multicolumn{5}{c|}{Trajectory} & \multicolumn{4}{c|}{Brightness} & Average \\ \midrule
Validation Dataset & Orig. & 1 & 2 & 3 & 4 & 5 & 6 & 7 & 8 & 9 & All \\ \midrule
No Adaptation & 48.43 & 45.17 & 36.58 & 42.28 & 26.57 & 38.70 & 24.39 & 32.76 & 38.99 & 37.37 & 35.89 \\
Mummadi \etal~\cite{tent_mum} & - & 45.85 & 45.73 & 46.25 & 42.39 & 44.13 & 27.82 & 36.48 & 45.19 & 44.85 & 42.08 \\
URIE~\cite{urie} & - & 40.88 & 41.04 & 42.03 & 37.68 & 40.17 & 20.14 & 30.18 & 41.44 & 40.14 & 37.08 \\
SENTRY~\cite{sentry} & - & 45.47 & 45.58 & 46.12 & 41.55 & 43.52 & 24.57 & 35.03 & 45.03 & 44.69 & 41.28 \\
Tent~\cite{tent} & - & 43.27 & 44.10 & 44.37 & 40.61 & 41.78 & 25.52 & 34.10 & 43.39 & 43.23 & 40.04 \\
Ev-TTA & - & \textbf{46.32} & \textbf{46.05} & \textbf{46.57} & \textbf{43.23} & \textbf{44.58} & \textbf{28.05} & \textbf{36.98} & \textbf{46.03} & \textbf{45.64} & \textbf{42.61} \\
\bottomrule
\end{tabular}
}
\caption{Online evaluation results of DiST~\cite{n_imagenet} on N-ImageNet~\cite{n_imagenet} and its variants.}
\label{tab:supp_full}
\end{table*}

\clearpage

{\small
\bibliographystyle{ieee_fullname}
\bibliography{egbib}

\begin{thebibliography}{10}\itemsep=-1pt

\bibitem{uda_2}
Cycada: Cycle consistent adversarial domain adaptation.
\newblock In {\em International Conference on Machine Learning (ICML)}, 2018.

\bibitem{ace}
I. {Alzugaray} and M. {Chli}.
\newblock Ace: An efficient asynchronous corner tracker for event cameras.
\newblock In {\em 2018 International Conference on 3D Vision (3DV)}, pages
  653--661, 2018.

\bibitem{tta_sr}
Michal~Irani Assaf~Shocher, Nadav~Cohen.
\newblock "zero-shot" super-resolution using deep internal learning.
\newblock In {\em The IEEE Conference on Computer Vision and Pattern
  Recognition (CVPR)}, June 2018.

\bibitem{tta_synth}
David Bau, Hendrik Strobelt, William Peebles, Jonas Wulff, Bolei Zhou, Jun-Yan
  Zhu, and Antonio Torralba.
\newblock Semantic photo manipulation with a generative image prior.
\newblock {\em ACM Trans. Graph.}, 38(4), July 2019.

\bibitem{ddd17}
Jonathan Binas, Daniel Neil, Shih-Chii Liu, and Tobi Delbruck.
\newblock Ddd17: End-to-end davis driving dataset, 2017.

\bibitem{matrix_lstm}
Marco Cannici, Marco Ciccone, Andrea Romanoni, and Matteo Matteucci.
\newblock A differentiable recurrent surface for asynchronous event-based data.
\newblock In {\em European Conference on Computer Vision (ECCV)}, August 2020.

\bibitem{binary_image_2}
G. {Cohen}, S. {Afshar}, G. {Orchard}, J. {Tapson}, R. {Benosman}, and A. {van
  Schaik}.
\newblock Spatial and temporal downsampling in event-based visual
  classification.
\newblock {\em IEEE Transactions on Neural Networks and Learning Systems},
  29(10):5030--5044, 2018.

\bibitem{v2e}
Tobi Delbruck, Yuhuang Hu, and Zhe He.
\newblock V2e: From video frames to realistic dvs event camera streams.
\newblock {\em arXiv preprint arXiv:2006.07722}, 2020.

\bibitem{amae}
Yongjian Deng, Youfu Li, and Hao Chen.
\newblock Amae: Adaptive motion-agnostic encoder for event-based object
  classification.
\newblock {\em IEEE Robotics and Automation Letters}, 5(3):4596--4603, 2020.

\bibitem{density_filter}
Yang Feng, Hengyi Lv, Hailong Liu, Yisa Zhang, Yuyao Xiao, and Chengshan Han.
\newblock Event density based denoising method for dynamic vision sensor.
\newblock {\em Applied Sciences}, 10(6), 2020.

\bibitem{uda_1}
Yaroslav Ganin and Victor Lempitsky.
\newblock Unsupervised domain adaptation by backpropagation.
\newblock In Francis Bach and David Blei, editors, {\em Proceedings of the 32nd
  International Conference on Machine Learning}, volume~37 of {\em Proceedings
  of Machine Learning Research}, pages 1180--1189, Lille, France, 07--09 Jul
  2015. PMLR.

\bibitem{cond_rat}
R.~C. Geary.
\newblock The frequency distribution of the quotient of two normal variates.
\newblock {\em Journal of the Royal Statistical Society}, 93(3):442--446, 1930.

\bibitem{vid2e}
Daniel Gehrig, Mathias Gehrig, Javier Hidalgo-Carri\'o, and Davide Scaramuzza.
\newblock Video to events: Recycling video datasets for event cameras.
\newblock In {\em {IEEE} Conf. Comput. Vis. Pattern Recog. (CVPR)}, June 2020.

\bibitem{est}
D. {Gehrig}, A. {Loquercio}, K. {Derpanis}, and D. {Scaramuzza}.
\newblock End-to-end learning of representations for asynchronous event-based
  data.
\newblock In {\em 2019 IEEE/CVF International Conference on Computer Vision
  (ICCV)}, pages 5632--5642, 2019.

\bibitem{tta_rl}
Nicklas Hansen, Rishabh Jangir, Yu Sun, Guillem Aleny{\`a}, Pieter Abbeel,
  Alexei~A Efros, Lerrel Pinto, and Xiaolong Wang.
\newblock Self-supervised policy adaptation during deployment.
\newblock In {\em International Conference on Learning Representations}, 2021.

\bibitem{resnet}
Kaiming He, Xiangyu Zhang, Shaoqing Ren, and Jian Sun.
\newblock Deep residual learning for image recognition.
\newblock pages 770--778, 06 2016.

\bibitem{ddd20}
Yuhuang Hu, Jonathan Binas, Daniel Neil, Shih-Chii Liu, and Tobi Delbruck.
\newblock Ddd20 end-to-end event camera driving dataset: Fusing frames and
  events with deep learning for improved steering prediction, 2020.

\bibitem{n_imagenet}
Junho Kim, Jaehyeok Bae, Gangin Park, Dongsu Zhang, and Young~Min Kim.
\newblock N-imagenet: Towards robust, fine-grained object recognition with
  event cameras.
\newblock In {\em Proceedings of the IEEE/CVF International Conference on
  Computer Vision (ICCV)}, pages 2146--2156, October 2021.

\bibitem{adam}
Diederik~P. Kingma and Jimmy Ba.
\newblock Adam: {A} method for stochastic optimization.
\newblock In Yoshua Bengio and Yann LeCun, editors, {\em 3rd International
  Conference on Learning Representations, {ICLR} 2015, San Diego, CA, USA, May
  7-9, 2015, Conference Track Proceedings}, 2015.

\bibitem{hots}
Xavier Lagorce, Garrick Orchard, Francesco Galluppi, Bert Shi, and Ryad
  Benosman.
\newblock Hots: A hierarchy of event-based time-surfaces for pattern
  recognition.
\newblock {\em IEEE transactions on pattern analysis and machine intelligence},
  39, 07 2016.

\bibitem{event_driving}
Ana~I. Maqueda, Antonio Loquercio, G. Gallego, N. Garc{\'i}a, and D.
  Scaramuzza.
\newblock Event-based vision meets deep learning on steering prediction for
  self-driving cars.
\newblock {\em 2018 IEEE/CVF Conference on Computer Vision and Pattern
  Recognition}, pages 5419--5427, 2018.

\bibitem{sim2real_gap_uda}
Nico Messikommer, Daniel Gehrig, Mathias Gehrig, and Davide Scaramuzza.
\newblock Bridging the gap between events and frames through unsupervised
  domain adaptation, 2021.

\bibitem{asynet}
Nico Messikommer, Daniel Gehrig, Antonio Loquercio, and Davide Scaramuzza.
\newblock Event-based asynchronous sparse convolutional networks.
\newblock 2020.

\bibitem{tent_mum}
Chaithanya~Kumar Mummadi, Robin Hutmacher, Kilian Rambach, Evgeny Levinkov,
  Thomas Brox, and Jan~Hendrik Metzen.
\newblock Test-time adaptation to distribution shift by confidence maximization
  and input transformation, 2021.

\bibitem{gauss}
D.A. Nix and A.S. Weigend.
\newblock Estimating the mean and variance of the target probability
  distribution.
\newblock In {\em Proceedings of 1994 IEEE International Conference on Neural
  Networks (ICNN'94)}, volume~1, pages 55--60 vol.1, 1994.

\bibitem{n_caltech}
Garrick Orchard, Ajinkya Jayawant, Gregory~K. Cohen, and Nitish Thakor.
\newblock Converting static image datasets to spiking neuromorphic datasets
  using saccades.
\newblock {\em Frontiers in Neuroscience}, 9:437, 2015.

\bibitem{timestamp_image}
P.~K.~J. {Park}, B.~H. {Cho}, J.~M. {Park}, K. {Lee}, H.~Y. {Kim}, H.~A.
  {Kang}, H.~G. {Lee}, J. {Woo}, Y. {Roh}, W.~J. {Lee}, C. {Shin}, Q. {Wang},
  and H. {Ryu}.
\newblock Performance improvement of deep learning based gesture recognition
  using spatiotemporal demosaicing technique.
\newblock In {\em 2016 IEEE International Conference on Image Processing
  (ICIP)}, pages 1624--1628, 2016.

\bibitem{pytorch}
Adam Paszke, Sam Gross, Francisco Massa, Adam Lerer, James Bradbury, Gregory
  Chanan, Trevor Killeen, Zeming Lin, Natalia Gimelshein, Luca Antiga, Alban
  Desmaison, Andreas Kopf, Edward Yang, Zachary DeVito, Martin Raison, Alykhan
  Tejani, Sasank Chilamkurthy, Benoit Steiner, Lu Fang, Junjie Bai, and Soumith
  Chintala.
\newblock Pytorch: An imperative style, high-performance deep learning library.
\newblock In H. Wallach, H. Larochelle, A. Beygelzimer, F. d\textquotesingle
  Alch\'{e}-Buc, E. Fox, and R. Garnett, editors, {\em Advances in Neural
  Information Processing Systems 32}, pages 8024--8035. Curran Associates,
  Inc., 2019.

\bibitem{megapixel}
Etienne Perot, Pierre de Tournemire, Davide Nitti, Jonathan Masci, and Amos
  Sironi.
\newblock Learning to detect objects with a 1 megapixel event camera.
\newblock {\em arXiv preprint arXiv:2009.13436}, 2020.

\bibitem{sentry}
Viraj Prabhu, Shivam Khare, Deeksha Kartik, and Judy Hoffman.
\newblock Sentry: Selective entropy optimization via committee consistency for
  unsupervised domain adaptation.
\newblock In {\em Proceedings of the IEEE/CVF International Conference on
  Computer Vision (ICCV)}, pages 8558--8567, October 2021.

\bibitem{esim}
Henri Rebecq, Daniel Gehrig, and Davide Scaramuzza.
\newblock {ESIM}: an open event camera simulator.
\newblock {\em Conf. on Robotics Learning (CoRL)}, Oct. 2018.

\bibitem{event2vid}
H. {Rebecq}, R. {Ranftl}, V. {Koltun}, and D. {Scaramuzza}.
\newblock Events-to-video: Bringing modern computer vision to event cameras.
\newblock In {\em 2019 IEEE/CVF Conference on Computer Vision and Pattern
  Recognition (CVPR)}, pages 3852--3861, 2019.

\bibitem{imagenet}
Olga Russakovsky, Jia Deng, Hao Su, Jonathan Krause, Sanjeev Satheesh, Sean Ma,
  Zhiheng Huang, Andrej Karpathy, Aditya Khosla, Michael Bernstein,
  Alexander~C. Berg, and Li Fei-Fei.
\newblock {ImageNet Large Scale Visual Recognition Challenge}.
\newblock {\em International Journal of Computer Vision (IJCV)},
  115(3):211--252, 2015.

\bibitem{uda_3}
Kate Saenko, Brian Kulis, Mario Fritz, and Trevor Darrell.
\newblock Adapting visual category models to new domains.
\newblock In Kostas Daniilidis, Petros Maragos, and Nikos Paragios, editors,
  {\em Computer Vision -- ECCV 2010}, pages 213--226, Berlin, Heidelberg, 2010.
  Springer Berlin Heidelberg.

\bibitem{snn}
Ali Samadzadeh, Fatemeh Sadat~Tabatabaei Far, Ali Javadi, Ahmad Nickabadi, and
  Morteza~Haghir Chehreghani.
\newblock Convolutional spiking neural networks for spatio-temporal feature
  extraction.
\newblock {\em arXiv preprint arXiv:2003.12346}, 2020.

\bibitem{eventnet}
Yusuke Sekikawa, Kosuke Hara, and Hideo Saito.
\newblock Eventnet: Asynchronous recursive event processing, 2019.

\bibitem{hats}
Amos Sironi, Manuele Brambilla, Nicolas Bourdis, Xavier Lagorce, and Ryad
  Benosman.
\newblock {HATS: Histograms of Averaged Time Surfaces for Robust Event-based
  Object Classification}.
\newblock {\em arXiv preprint arXiv:2018.00186}, June 2018.

\bibitem{gen3}
B. {Son}, Y. {Suh}, S. {Kim}, H. {Jung}, J. {Kim}, C. {Shin}, K. {Park}, K.
  {Lee}, J. {Park}, J. {Woo}, Y. {Roh}, H. {Lee}, Y. {Wang}, I. {Ovsiannikov},
  and H. {Ryu}.
\newblock 4.1 a 640×480 dynamic vision sensor with a 9µm pixel and 300meps
  address-event representation.
\newblock In {\em 2017 IEEE International Solid-State Circuits Conference
  (ISSCC)}, pages 66--67, 2017.

\bibitem{urie}
Taeyoung Son, Juwon Kang, Namyup Kim, Sunghyun Cho, and Suha Kwak.
\newblock Urie: Universal image enhancement for visual recognition in the wild.
\newblock In {\em ECCV}, 2020.

\bibitem{reduce_sim2real}
T. Stoffregen, C. Scheerlinck, D. Scaramuzza, T. Drummond, N. Barnes, L.
  Kleeman, and R. Mahoney.
\newblock Reducing the sim-to-real gap for event cameras.
\newblock In {\em European Conference on Computer Vision (ECCV)}, august 2020.

\bibitem{ttt}
Yu Sun, Xiaolong Wang, Zhuang Liu, John Miller, Alexei Efros, and Moritz Hardt.
\newblock Test-time training with self-supervision for generalization under
  distribution shifts.
\newblock In Hal~Daumé III and Aarti Singh, editors, {\em Proceedings of the
  37th International Conference on Machine Learning}, volume 119 of {\em
  Proceedings of Machine Learning Research}, pages 9229--9248. PMLR, 13--18 Jul
  2020.

\bibitem{timelens}
Stepan Tulyakov, Daniel Gehrig, Stamatios Georgoulis, Julius Erbach, Mathias
  Gehrig, Yuanyou Li, and Davide Scaramuzza.
\newblock {TimeLens}: Event-based video frame interpolation.
\newblock {\em IEEE Conference on Computer Vision and Pattern Recognition},
  2021.

\bibitem{uda_4}
Eric Tzeng, Judy Hoffman, Ning Zhang, Kate Saenko, and Trevor Darrell.
\newblock Deep domain confusion: Maximizing for domain invariance.
\newblock {\em CoRR}, abs/1412.3474, 2014.

\bibitem{tsne}
Laurens van~der Maaten and Geoffrey Hinton.
\newblock Visualizing data using {t-SNE}.
\newblock {\em Journal of Machine Learning Research}, 9:2579--2605, 2008.

\bibitem{mathstat}
Dennis~D. Wackerly, William~Mendenhall III, and Richard~L. Scheaffer.
\newblock {\em Mathematical Statistics with Applications}.
\newblock Duxbury Advanced Series, sixth edition edition, 2002.

\bibitem{tent}
Dequan Wang, Evan Shelhamer, Shaoteng Liu, Bruno Olshausen, and Trevor Darrell.
\newblock Tent: Fully test-time adaptation by entropy minimization.
\newblock In {\em International Conference on Learning Representations}, 2021.

\bibitem{ev_gait}
Y. {Wang}, B. {Du}, Y. {Shen}, K. {Wu}, G. {Zhao}, J. {Sun}, and H. {Wen}.
\newblock Ev-gait: Event-based robust gait recognition using dynamic vision
  sensors.
\newblock In {\em 2019 IEEE/CVF Conference on Computer Vision and Pattern
  Recognition (CVPR)}, pages 6351--6360, 2019.

\bibitem{denoise_exp_special}
J. {Wu}, C. {Ma}, X. {Yu}, and G. {Shi}.
\newblock Denoising of event-based sensors with spatial-temporal correlation.
\newblock In {\em ICASSP 2020 - 2020 IEEE International Conference on
  Acoustics, Speech and Signal Processing (ICASSP)}, pages 4437--4441, 2020.

\bibitem{evflownet}
Alex Zhu, Liangzhe Yuan, Kenneth Chaney, and Kostas Daniilidis.
\newblock Ev-flownet: Self-supervised optical flow estimation for event-based
  cameras.
\newblock In {\em Proceedings of Robotics: Science and Systems}, Pittsburgh,
  Pennsylvania, June 2018.

\end{thebibliography}
}

\end{document}